\documentclass[10pt,twocolumn,letterpaper]{article}

\usepackage{cvpr}
\usepackage{times}
\usepackage{epsfig}
\usepackage{graphicx}
\usepackage{amsmath}
\usepackage{amssymb}

\usepackage{subcaption}
\usepackage{float}
\usepackage{sidecap}
\usepackage{booktabs}

\usepackage[breaklinks=true,bookmarks=false]{hyperref}

\cvprfinalcopy 

\def\cvprPaperID{52} 

\begin{document}
\definecolor{colour}{RGB}{228, 24, 133}
\title{ManiGAN: Text-Guided Image Manipulation}

\title{ManiGAN: Text-Guided Image Manipulation}

\author{Bowen Li$^{1}$ \quad Xiaojuan Qi$^{1,2}$ \quad Thomas Lukasiewicz$^{1}$ \quad Philip H. S. Torr$^{1}$\\
$^{1}$University of Oxford \quad $^{2}$University of Hong Kong\\
{\tt\small \{bowen.li, thomas.lukasiewicz\}@cs.ox.ac.uk}  \quad
{\tt\small \{xiaojuan.qi, philip.torr\}@eng.ox.ac.uk}
}

\maketitle

\begin{abstract}

The goal of our paper is to semantically edit parts of an image to match a given text that describes desired attributes (e.g., texture, colour, and background), while preserving other contents that are irrelevant to the text. To achieve this, we propose a novel generative adversarial network (ManiGAN), which contains two key components: text-image affine combination module (ACM) and detail correction module (DCM).
The ACM selects image regions relevant to the given text and then correlates the regions with corresponding semantic words for effective manipulation. Meanwhile, it encodes original image features to help reconstruct text-irrelevant contents.
The DCM rectifies mismatched attributes and {completes missing contents} of the synthetic image.
Finally, we suggest a new metric for evaluating image manipulation results, in terms of both the generation of new attributes and 
the reconstruction of text-irrelevant contents. Extensive experiments on the CUB and COCO datasets demonstrate the superior performance of the proposed method. Code is available at \textcolor{colour}{https://github.com/mrlibw/ManiGAN}. 

\end{abstract}
\vspace{-3.5mm}
\section{Introduction}
Image manipulation aims to modify some aspects of given images, from low-level colour or texture ~\cite{gatys2016image, zhang2016colorful} to high level semantics~\cite{zhu2016generative}, to meet a user's preferences, which has numerous potential applications in video games, image editing, and computer-aided design. Recently, with the development of deep learning and deep generative models, automatic image manipulation has made remarkable progress, including image inpainting \cite{iizuka2016let, pathak2016context}, image colourisation \cite{zhang2016colorful}, style transfer \cite{gatys2016image, johnson2016perceptual}, and domain or attribute translation \cite{isola2017image, lample2017fader}.

\begin{figure*}[t]
\begin{minipage}{1\textwidth}
\includegraphics[width=1\linewidth, height=0.41\linewidth]{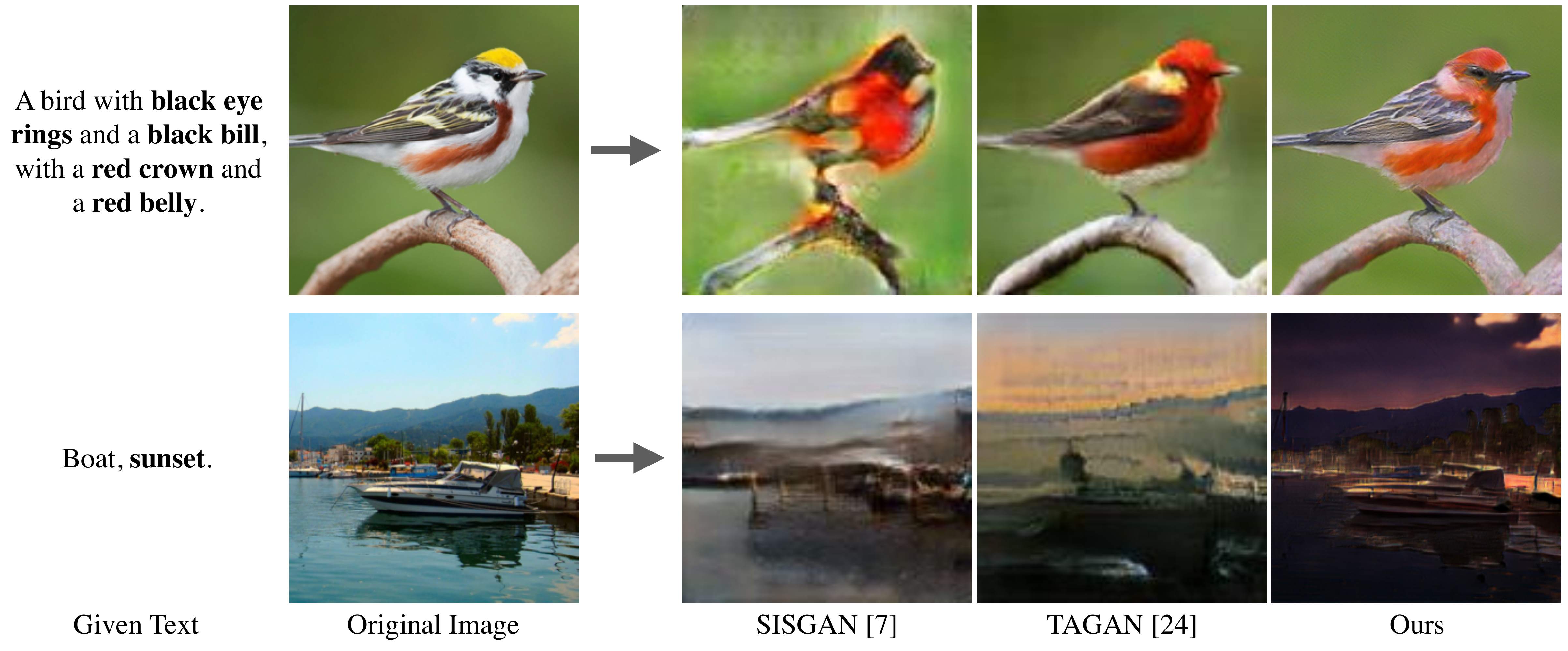}
\end{minipage}

\centering
\caption{Given an original image that needs to be edited and a text provided by a user describing desired attributes, the goal is to edit parts of the image according to the given text while preserving text-irrelevant contents. Current state-of-the-art methods only generate low-quality images, and fail to do manipulation on COCO. In contrast, our method allows the original image to be manipulated accurately to match the given description, and also reconstructs text-irrelevant contents.}
\label{fig:qual_show}
\end{figure*}

All the above works mainly focus on specific problems, and few studies \cite{dong2017semantic,nam2018text} concentrate on more general and user-friendly image manipulation by using natural language descriptions. 
More precisely, the task aims to semantically edit parts of an image according to the given text provided by a user, while preserving other contents that are not described in the text. However, current state-of-the-art text-guided image manipulation methods are only able to produce low-quality images (see Fig.~\ref{fig:qual_show}: first row), far from satisfactory, and even fail to effectively manipulate complex scenes (see Fig.~\ref{fig:qual_show}: second row).

To achieve effective image manipulation guided by text descriptions, the key is to exploit both text and image cross-modality information, generating new attributes matching the given text and also preserving text-irrelevant contents of the original image.
To fuse text and image information, existing methods~\cite{dong2017semantic,nam2018text} typically choose to directly concatenate image and global sentence features along the channel direction.
Albeit simple, the above heuristic may suffer from some potential issues. 
Firstly, the model cannot precisely correlate fine-grained words with corresponding visual attributes that need to be modified, leading to inaccurate and coarse modification. For instance, shown in the first row of Fig.~\ref{fig:qual_show}, both models cannot generate detailed visual attributes like \textit{black eye rings} and a \textit{black bill}.
Secondly, the model cannot effectively identify text-irrelevant contents and thus fails to reconstruct them, resulting in undesirable modification of text-irrelevant parts in the image. For example, in Fig.~\ref{fig:qual_show}, besides modifying the required attributes, both models~\cite{dong2017semantic,nam2018text} also change the texture of the bird (first row) and the structure of the scene (second row).

To address the above issues, we propose a novel generative adversarial network for text-guided image manipulation (ManiGAN), which can generate high-quality new attributes matching the given text, and at the same time effectively reconstruct text-irrelevant contents of the original image.
The key is a text-image affine combination module (ACM) where text and image features collaborate to select text-relevant regions that need to be modified, and then correlate those regions with corresponding semantic words for generating new visual attributes semantically aligned with the given text description. Meanwhile, it also encodes original image representations for reconstructing text-irrelevant contents.
Besides, to further enhance the results, we introduce a detail correction module (DCM) which can rectify mismatched attributes and {complete missing contents}. Our final model can produce high-quality manipulation results with fine-grained details (see Fig.~\ref{fig:qual_show}: Ours).

Finally, we suggest a new metric to assess image manipulation results. The metric can appropriately reflect the performance of image manipulation, in terms of both the generation of new visual attributes corresponding to the given text, and the reconstruction of text-irrelevant contents of the original image. Extensive experiments on the CUB \cite{wah2011caltech} and COCO \cite{lin2014microsoft} datasets demonstrate the superiority of our model, where our model outperforms existing state-of-the-art methods both qualitatively and quantitatively.

\section{Related Work}

\noindent\textbf{Text-to-image generation} has drawn much attention due to the success of GANs \cite{goodfellow2014generative} in generating realistic images. Reed et al.~\cite{reed2016generative} proposed to use conditional GANs to generate plausible images from given text descriptions. Zhang et al.~\cite{zhang2017stackgan, zhang2018stackgan++} stacked multiple GANs to generate high-resolution images from coarse- to fine-scale. Xu et at.~\cite{xu2018attngan} and Li et al.~\cite{li2019controllable} implemented attention mechanisms to explore fine-grained information at the word-level. However, all aforementioned methods mainly focus on generating new photo-realistic images from texts, and not on manipulating specific visual attributes of given images using natural language descriptions. 

\smallskip\noindent\textbf{Conditional image synthesis.} Our work is related to conditional image synthesis \cite{brock2016neural, chen2018language, cheng2018sequential, el2019tell, liu2017unsupervised, mo2018instagan, park2019semantic, tang2019multi, tang2019local, zhu2016generative}. Recently, various methods have been proposed to achieve paired image-to-image translation \cite{chen2017photographic, isola2017image, wang2018high}, or unpaired translation \cite{liu2016coupled, taigman2016unsupervised,zhu2017unpaired}. However, all these methods mainly focus on same-domain image translation instead of image manipulation using cross-domain text descriptions.

\smallskip\noindent\textbf{Text-guided image manipulation.} There are few studies focusing on image manipulation using natural language descriptions. Dong et al.~\cite{dong2017semantic} proposed a GAN-based encoder-decoder architecture to disentangle the semantics of both input images and text descriptions. Nam et al.~\cite{nam2018text} implemented a similar architecture, but introduced a text-adaptive discriminator that can provide specific word-level training feedback to the generator. However, both methods are limited in performance due to a less effective text-image concatenation method and a coarse sentence condition. 

\smallskip\noindent\textbf{Affine transformation} has been widely implemented in conditional normalisation techniques~\cite{de2017modulating, dumoulin2016learned, huang2017arbitrary, miyato2018cgans, park2019semantic, perez2018film} to incorporate additional information~\cite{dumoulin2016learned, huang2017arbitrary, miyato2018cgans}, or to avoid information loss caused by normalisation~\cite{park2019semantic}. Differently from these methods, our affine combination module is designed to 
fuse text and image cross-modality representations to enable effective manipulation, and is only placed at specific positions instead of all normalisation layers. 


\section{Generative Adversarial Networks for Image Manipulation}

\begin{figure*}[t]
\begin{minipage}{0.943\textwidth}
\includegraphics[width=1\linewidth, height=0.31\linewidth]{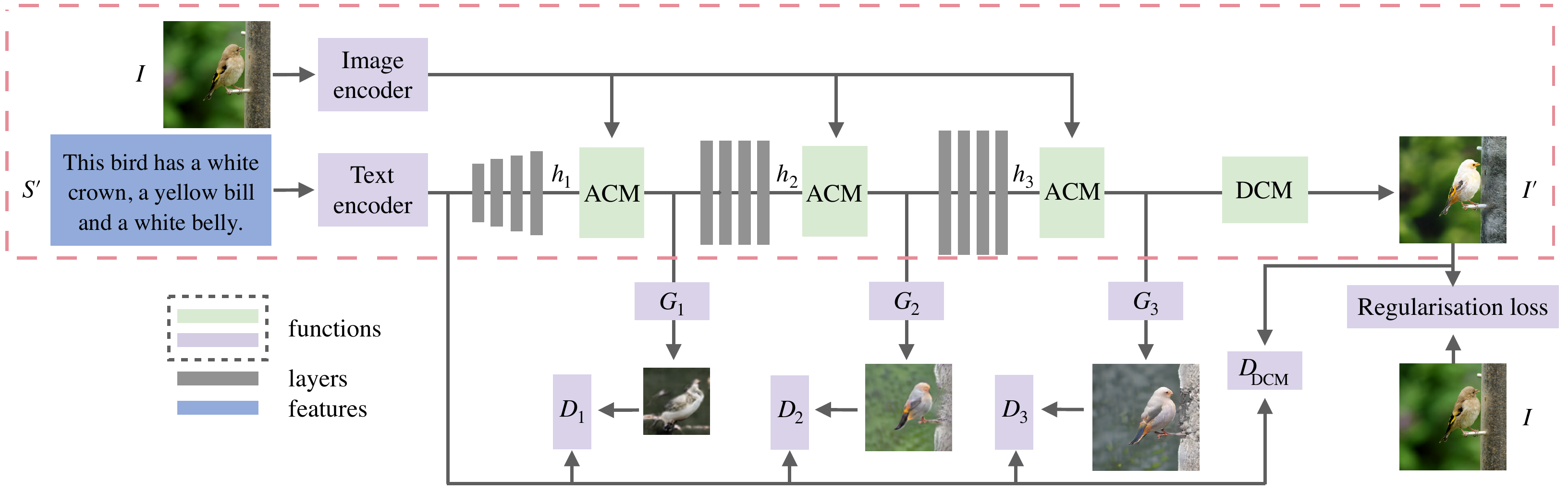}
\end{minipage}
\caption{The architecture of ManiGAN. The red dashed box indicates the inference pipeline that a text description $S'$ is given by a user, while in training, the text $S'$ is replaced by $S$ that correctly describes $I$. ACM denotes the text-image affine combination module. DCM denotes the detail correction module. The attention is omitted for simplicity. Please see supplementary material for full architecture.}
\label{fig:archi}
\end{figure*}

Given an input image $I$, and a text description ${S}'$ provided by a user, the model aims to generate a manipulated image $I'$ that is semantically aligned with ${S}'$ while preserving text-irrelevant contents existing in $I$.
To achieve this, we propose two novel components: (1) a text-image affine combination module (ACM), and (2) a detail correction module (DCM). We elaborate our model as follows. 

\subsection{Architecture}
As shown in Fig.~\ref{fig:archi}, we adopt the multi-stage ControlGAN \cite{li2019controllable} architecture as the basic framework, as it achieves high-quality and controllable image generation results based on the given text descriptions. 
We add an image encoder, which is a pretrained Inception-v3 network~\cite{szegedy2016rethinking}, to extract regional image representations {$v$}.
Our proposed text-image affine combination module (ACM) is utilised to fuse text representations, encoded from a pretrained RNN \cite{nam2018text}, and regional image representations before each upsampling block at the end of each stage. 
For each stage, the text features are refined with several convolutional layers to produce hidden features $h$. 
{The proposed ACM further combines $h$ with the original image features $v$ in order to effectively select image regions corresponding to the given text, and then correlate those regions with text information for accurate manipulation. Meanwhile, it also encodes the original image representations for stable reconstruction.}
The output features from the ACM module are fed into the corresponding generator to produce an edited image, and are also upsampled serving as input to the next stage for image manipulation at a higher resolution. 
The whole framework gradually generates new visual attributes matching the given text description at a higher resolution with higher quality, and also reconstructs text-irrelevant contents existing in the input image at a finer scale. Finally, the proposed detail correction module (DCM) is used to rectify inappropriate attributes, and to complete missing details. 
\begin{figure*}[t]
\centering
\begin{minipage}{0.989\textwidth}
\includegraphics[width=1\linewidth, height=0.369\linewidth]{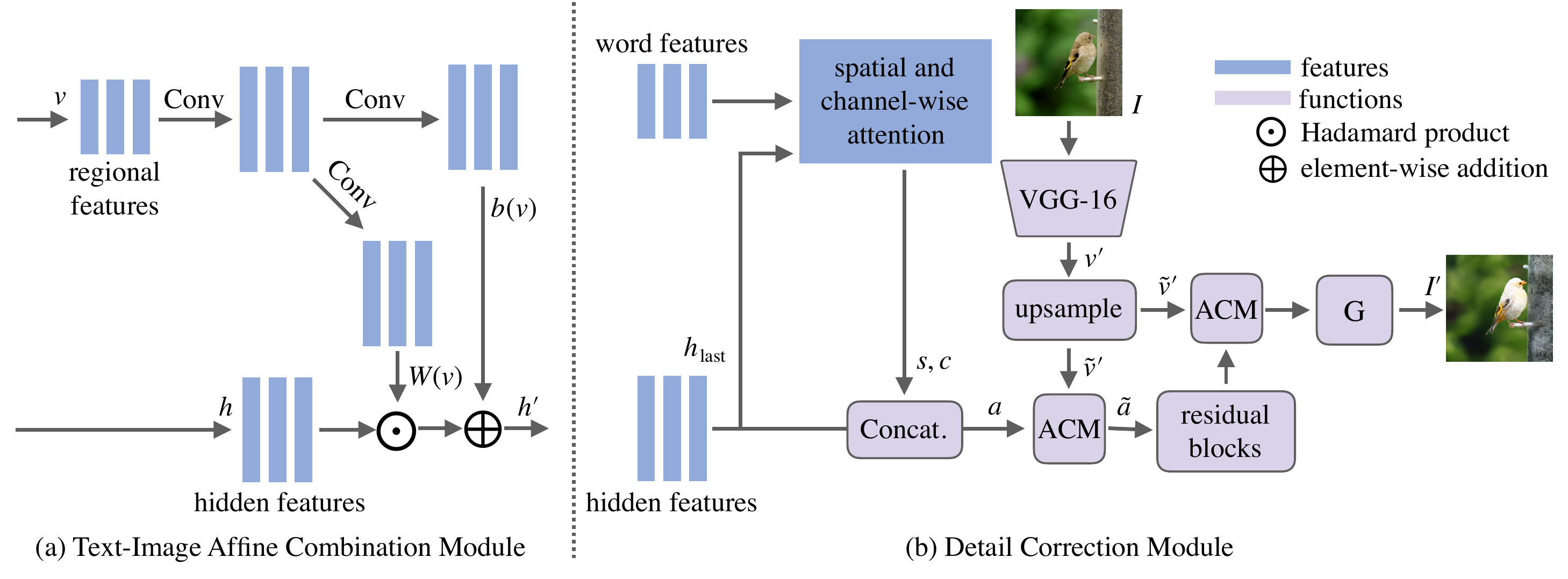}
\end{minipage}
\caption{The architecture of the text-image affine combination module and the detail correction module. In (b), ACM denotes the text-image affine combination module.}
\label{fig:archi_modules}
\end{figure*}
\subsection{Text-Image Affine Combination Module}
{The existing concatenation scheme for combining text-image cross-modality representations cannot effectively locate desired regions that need to be modified, and thus fails to achieve fine-grained image manipulation, regarding both the generation quality of new attributes corresponding to the given text, and the reconstruction stability of text-irrelevant image contents. To address the above issue, we propose a simple text-image affine combination module to fuse text-image cross-modality representations as discussed below.}

As shown in Fig.~\ref{fig:archi_modules}~(a), our affine combination module takes two inputs: (1) the hidden features $h \in \mathbb{R}^{C\times H \times D}$ from the input text or intermediate hidden representations between two stages, where $C$ is the number of channels, $H$ and $D$ are the height and width of the feature map, respectively, and (2) the regional image features $v \in \mathbb{R}^{256 \times 17 \times 17}$ from the input image $I$ encoded by the Inception-v3 network \cite{szegedy2016rethinking}.
Then, $v$ is upsampled and further processed with two convolutional layers to produce $W(v)$ and $b(v)$ that have the same size as $h$.
Finally, we fuse the two modality representations to produce  ${h}' \in \mathbb{R}^{C \times H \times D}$ as
\begin{equation}
{h}'=h \odot W(v) + b(v),
\end{equation}
where $W(v)$ and $b(v)$ are the learned weights and biases based on the regional image features $v$, and $\odot$ denotes Hadamard element-wise product. 
We use $W$ and $b$ to represent the functions that convert the regional features $v$ to scaling and bias values.

Our affine combination module (ACM) is designed to fuse text and image cross-modality representations. $W(v)$ and $b(v)$ encode the input image into semantically meaningful features as shown in Fig.~\ref{fig:ana_vis}.
The multiplication operation enables text representations $h$ to re-weight image feature maps, which serves as a regional selection purpose to 
{help the model precisely identify desired attributes matching the given text, and in the meantime the correlation between attributes and semantic words is built for effectively manipulation}. The bias term encodes image information to help the model stably reconstruct text-irrelevant contents.
The above is in contrast with previous approaches~\cite{de2017modulating, dumoulin2016learned, huang2017arbitrary, park2019semantic} which apply conditional affine transformation in normalisation layers to compensate potential information loss due to normalisation~\cite{park2019semantic} or to incorporate style information from a style image~\cite{dumoulin2016learned,huang2017arbitrary}.
To better understand what has been actually learned by different components of our affine combination module, we give a deeper analysis in Sec.~\ref{sec:ablation}.

\noindent\textbf{Why does the affine combination module work better than concatenation?}
By simply concatenating the text and image representations along the channel direction, existing models cannot explicitly distinguish regions that are required to be modified or to be reconstructed, which makes it hard to achieve a good balance between the generation of new attributes and the reconstruction of original contents. As a result, this imbalance leads to either inaccurate/coarse modification or changing text-irrelevant contents.
In contrast, our affine combination module uses multiplication on text and image representations to achieve a regional selection effect, aiding the model to focus on generating required fine-grained visual attributes.
Besides, the additive bias part encodes text-irrelevant image information 
to help reconstruct contents that are not required to be edited.


\subsection{Detail Correction Module}
\label{sec:dcm}
To further enhance the details and complete missing contents {in the synthetic image}, we propose a detail correction module (DCM), exploiting word-level text information and fine-grained image features.

As shown in Fig.~\ref{fig:archi_modules}~(b), our detail correction module takes three inputs: (1) the last hidden features $h_\text{last} \in \mathbb{R}^{{C}' \times H' \times D'}$ from the last affine combination module, (2) the word features encoded by a pretrained RNN following~\cite{xu2018attngan}, where each word is associated with a feature vector, and (3) visual features ${v}' \in \mathbb{R}^{128 \times 128 \times 128}$ that are extracted from the input image $I$, 
which are the \emph{relu2\_2} layer representations from a pretrained VGG-16 \cite{simonyan2014very} network.


Firstly, to further incorporate fine-grained word-level representations into hidden features $h_\text{last}$, we adopt the spatial attention and channel-wise attention introduced in \cite{li2019controllable} to generate spatial and channel-wise attention features $s \in \mathbb{R}^{C' \times H' \times D'}$ and $c \in \mathbb{R}^{C' \times H' \times D'}$, respectively, which are further concatenated with $h_\text{last}$ to produce intermediate features $a$. The features $a$ can further aid the model to refine visual attributes that are relevant to the given text, contributing to a more accurate and effective modification of the contents corresponding to the given description.
Secondly, to introduce detailed visual features from the input image for high-quality reconstruction, the shallow representations $v'$ of layer \emph{relu2\_2} from the pretrained VGG network are utilised, which are further upsampled to be the same size as $a$, denoted as $\tilde{v}'$.
Then, our proposed affine attention module is utilised to fuse visual representations $\tilde{v}'$ and hidden representations $a$, producing features $\tilde{a}$.
Finally, we refine $\tilde{a}$ with two residual blocks (details in the supplementary material) to generate the final manipulated image $I'$. 
\noindent\textbf{Why does the detail correction module work?} 
This module aims to refine the manipulated results by enhancing details and completing missing contents. On the one hand, the word-level spatial and channel-wise attentions closely correlate fine-grained word-level information with the intermediate feature maps, enhancing the detailed attribute modification. On the other hand, the shallow neural network layer is adopted to derive visual representations, which contain more detailed colour, texture, and edge information, contributing to missing detail construction.
Finally, further benefiting from our ACM, the above fine-grained text-image representations collaborate to enhance the quality.


\subsection{Training}
\label{sec:objective}
To train the network, 
we follow~\cite{li2019controllable} and adopt adversarial training, where our network and the discriminators ($D_1$, $D_2$, $D_3$, $D_\text{DCM}$) are alternatively optimised. 
Please see supplementary material for more details about training objectives. We only highlight some training differences compared with~\cite{li2019controllable}.

\smallskip\noindent\textbf{Generator objective.}
We follow the ControlGAN~\cite{li2019controllable} to construct the objective function for training the generator.
Besides, we add a regularisation term as
\begin{equation}
\mathcal{L}_\text{reg}=1-\frac{1}{CHW}|| {I}'-I ||,
\label{equ:reg}
\end{equation}
where $I$ is the real image sampled from the true image distribution, and $I'$ is the corresponding modified result produced by our model. 
The regularisation term is used to ensure diversity and to prevent the network learning an identity mapping, since this term can produce a large penalty when the generated image $I'$ is the same as the input image.

\smallskip\noindent\textbf{Discriminator objective.}
The loss function for the discriminator follows those used in ControlGAN \cite{li2019controllable}, and the function used to train the discriminator in the detail correction module is the same as the one used in the last stage of the main module.

\smallskip\noindent\textbf{Training.} 
Differently from~\cite{li2019controllable}, which has paired sentence $S$ and corresponding ground-truth image $I$ for training text-guided image generation models to learn the mapping $S$ $\rightarrow$ $I$, existing datasets such as COCO~\cite{lin2014microsoft} and CUB~\cite{wah2011caltech} with natural language descriptions do not provide paired training data ($I$, $S'$) $\rightarrow$ $I'_\text{gt}$ for training text-guided image manipulation models, where $S'$ is a text describing new attributes, and $I'_\text{gt}$ is the corresponding ground truth modified image.
 
To simulate the training data, we use paired data ($I$, $S$) $\rightarrow$ $I$ to train the model, and adopt $S'$ to construct the loss function following~\cite{li2019controllable}.
A natural question may arise: how does the model learn to modify the image $I$ if the input image $I$ and ground-truth image are the same, and the modified sentence $S'$ does not exist in the input? In theory, the optimal solution is that the network becomes an identity mapping from the input image to the output. 
The text-guided image manipulation model is required to jointly solve image generation from text descriptions ($S$ $\rightarrow$ $I$), similarly to~\cite{li2019controllable}, and text-irrelevant contents reconstruction ($I$ $\rightarrow$ $I$). 
Thanks to our proposed affine combination module, our model gains the capacity to disentangle regions required to be edited and regions needed to be preserved. Also, to generate new contents semantically matching the given text, the paired data $S$ and $I$ can serve as explicit supervision.

Moreover, to prevent the model from learning an identity mapping and to promote the model to learn a good ($S$ $\rightarrow$ $I$) mapping in the regions relevant to the given text, we propose the following training schemes.
Firstly, we introduce a regularisation term $\mathcal{L}_\text{reg}$ as Eq.~\eqref{equ:reg} in the generator objective to produce a penalty if the generated image becomes the same as the input image.
Secondly, we choose to early stop the training when the model achieves the best trade-off between the generation of new visual attributes aligned with the given text descriptions and the reconstruction of text-irrelevant contents existing in the original images.
The stop criterion is determined by evaluating the model on a hold-out validation and  measuring the results by our proposed image manipulation evaluation metric, called manipulative precision (see Fig.~\ref{fig:epochs}), which is discussed in Sec.~\ref{sec:experiment}.

\noindent\begin{figure*}[t]
\noindent\begin{minipage}{1\textwidth}
\scriptsize{The bird has a \textbf{black bill}, a \textbf{red crown}, and a \textbf{white belly}. (top)}\\
\scriptsize{This bird has \textbf{wings} that are \textbf{black}, and has a \textbf{red belly} and a \textbf{red head}. (bottom)}
\end{minipage}

\noindent\begin{minipage}{0.138\textwidth}
\includegraphics[width=1\linewidth, height=1\linewidth]{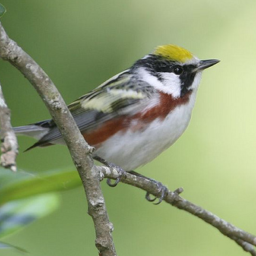}
\end{minipage}
\noindent\begin{minipage}{0.138\textwidth}
\includegraphics[width=1\linewidth, height=1\linewidth]{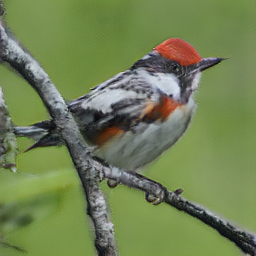}
\end{minipage}
\begin{minipage}{0.138\textwidth}
\includegraphics[width=1\linewidth, height=1\linewidth]{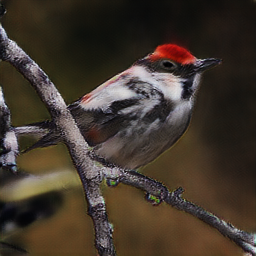}
\end{minipage}
\noindent\begin{minipage}{0.138\textwidth}
\includegraphics[width=1\linewidth, height=1\linewidth]{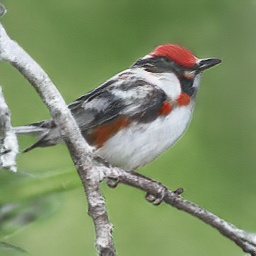}
\end{minipage}
\noindent\begin{minipage}{0.138\textwidth}
\includegraphics[width=1\linewidth, height=1\linewidth]{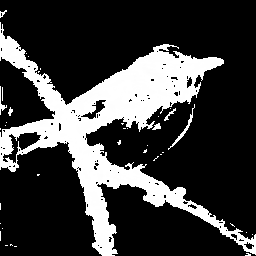}
\end{minipage}
\noindent\begin{minipage}{0.138\textwidth}
\includegraphics[width=1\linewidth, height=1\linewidth]{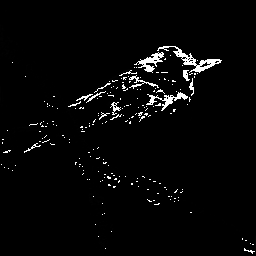}
\end{minipage}
\noindent\begin{minipage}{0.138\textwidth}
\includegraphics[width=1\linewidth, height=1\linewidth]{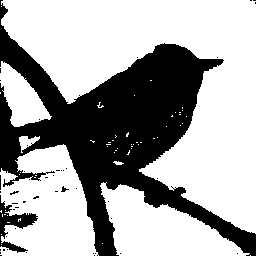}
\end{minipage}
\smallskip

\noindent\begin{minipage}{0.138\textwidth}
\includegraphics[width=1\linewidth, height=1\linewidth]{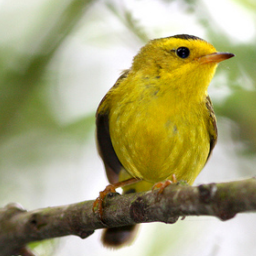}
\end{minipage}
\noindent\begin{minipage}{0.138\textwidth}
\includegraphics[width=1\linewidth, height=1\linewidth]{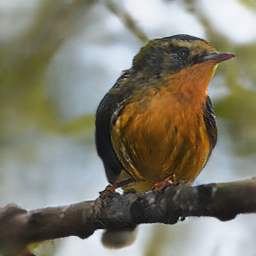}
\end{minipage}
\begin{minipage}{0.138\textwidth}
\includegraphics[width=1\linewidth, height=1\linewidth]{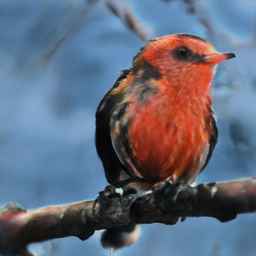}
\end{minipage}
\noindent\begin{minipage}{0.138\textwidth}
\includegraphics[width=1\linewidth, height=1\linewidth]{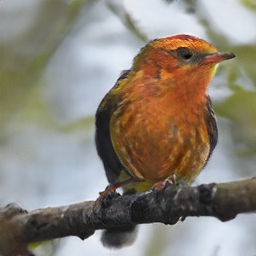}
\end{minipage}
\noindent\begin{minipage}{0.138\textwidth}
\includegraphics[width=1\linewidth, height=1\linewidth]{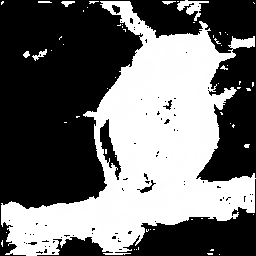}
\end{minipage}
\noindent\begin{minipage}{0.138\textwidth}
\includegraphics[width=1\linewidth, height=1\linewidth]{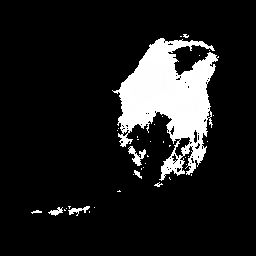}
\end{minipage}
\noindent\begin{minipage}{0.138\textwidth}
\includegraphics[width=1\linewidth, height=1\linewidth]{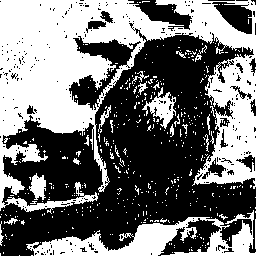}
\end{minipage}
\smallskip

\noindent\begin{minipage}{0.138\textwidth}
\centering
\scriptsize{Original}
\end{minipage}
\noindent\begin{minipage}{0.138\textwidth}
\centering
\scriptsize{Ours w/o W}
\end{minipage}
\begin{minipage}{0.138\textwidth}
\centering
\scriptsize{Ours w/o b}
\end{minipage}
\noindent\begin{minipage}{0.138\textwidth}
\centering
\scriptsize{Ours}
\end{minipage}
\noindent\begin{minipage}{0.138\textwidth}
\centering
\scriptsize{$W(v)$}
\end{minipage}
\noindent\begin{minipage}{0.138\textwidth}
\centering
\scriptsize{$h \odot W(v)$}
\end{minipage}
\noindent\begin{minipage}{0.138\textwidth}
\centering
\scriptsize{$b(v)$}
\end{minipage}

\centering
\caption{Ablation studies of the learned $W$ and $b$. The texts on the top are the given descriptions containing desired visual attributes, and the last three columns are the channel feature maps of $W(v)$, $h \odot W(v)$, and $b(v)$.}
\label{fig:ana_vis}
\end{figure*}

\section{Experiments}
\label{sec:experiment}
\begin{table}[h!]
  \centering
  \scalebox{0.75}{
  \begin{tabular}{c||cccc|cccc}
    \toprule
    \multicolumn{1}{c}{} &
    \multicolumn{4}{c}{CUB} & 
    \multicolumn{4}{c}{COCO}     \\             
    \cmidrule(r){2-5}
    \cmidrule(r){6-9}
    Method & IS  & sim  & diff  & MP & IS & sim & diff & MP \\
    \midrule
    SISGAN~\cite{dong2017semantic} & 2.24 & .045 & .508 & .022 & 3.44 & .077 & .442 & .042 \\
    TAGAN~\cite{nam2018text}  & 3.32 & .048 & .267 & .035 & 3.28 & .089 & .545 & .040 \\
    Ours w/o ACM  &  4.01  & \textbf{.138}  & .491 & .070 & 5.26 & .121 & .537 & .056\\    
    Ours w/ Concat.   & 3.81 & .135 & .512 & .065 & 13.48 & .085 & .532 & .039\\
    Ours w/o main   & \textbf{8.48} & .084 & \textbf{.235} & .064 & \textbf{17.59} & .080 & \textbf{.169} & .066\\
    Ours w/o DCM   & 3.84    & .123 & .447 & .068 & 6.99 & \textbf{.138} & .517 & .066\\
    \textbf{Ours}   & 8.47   & .101 & .281 & \textbf{.072} & 14.96 & .087 & .216 & \textbf{.068}\\
    \bottomrule
  \end{tabular}
  }
  \caption{Quantitative comparison: inception score (IS), text-image similarity (sim), $L_{1}$ pixel difference (diff), and manipulative precision (MP) of state-of-the-art approaches and ManiGAN on the CUB and COCO datasets. ``w/o ACM" denotes without the affine combination module. ``w/ Concat." denotes using concatenation method to combine hidden and image features. ``w/o main" denotes without main module. ``w/o DCM" denotes without detail correction module. For IS, similarity, and MP, higher is better; for pixel difference, lower is better.}
\label{table:quantitative}
\end{table}

Our model is evaluated on the CUB bird \cite{wah2011caltech} and more complicated COCO \cite{lin2014microsoft} datasets, comparing with two state-of-the-art approaches SISGAN \cite{dong2017semantic} and TAGAN \cite{nam2018text} on image manipulation using natural language descriptions.


\smallskip\noindent\textbf{Datasets.}\label{sec:datasets}
CUB bird \cite{wah2011caltech}: there are 8,855 training images and 2,933 test images, and each image has 10 corresponding text descriptions. COCO \cite{lin2014microsoft}: there are 82,783 training images and 40,504 validation images, and each image has 5 corresponding text descriptions. We preprocess these two datasets according to the method in \cite{xu2018attngan}.  

\smallskip\noindent\textbf{Implementation.}
In our setting, we train the detail correction module (DCM) separately from the main module. Once the main module has converged, we train the DCM subsequently and set the main module as the eval mode. There are three stages in the main module, and each stage contains a generator and a discriminator. We train three stages at the same time, and three different-scale images \(64 \times 64, 128 \times 128, 256 \times 256\) are generated progressively. 

\begin{figure}[t]
\centering
\begin{minipage}{0.45\textwidth}
\includegraphics[width=1\linewidth, height=0.44\linewidth]{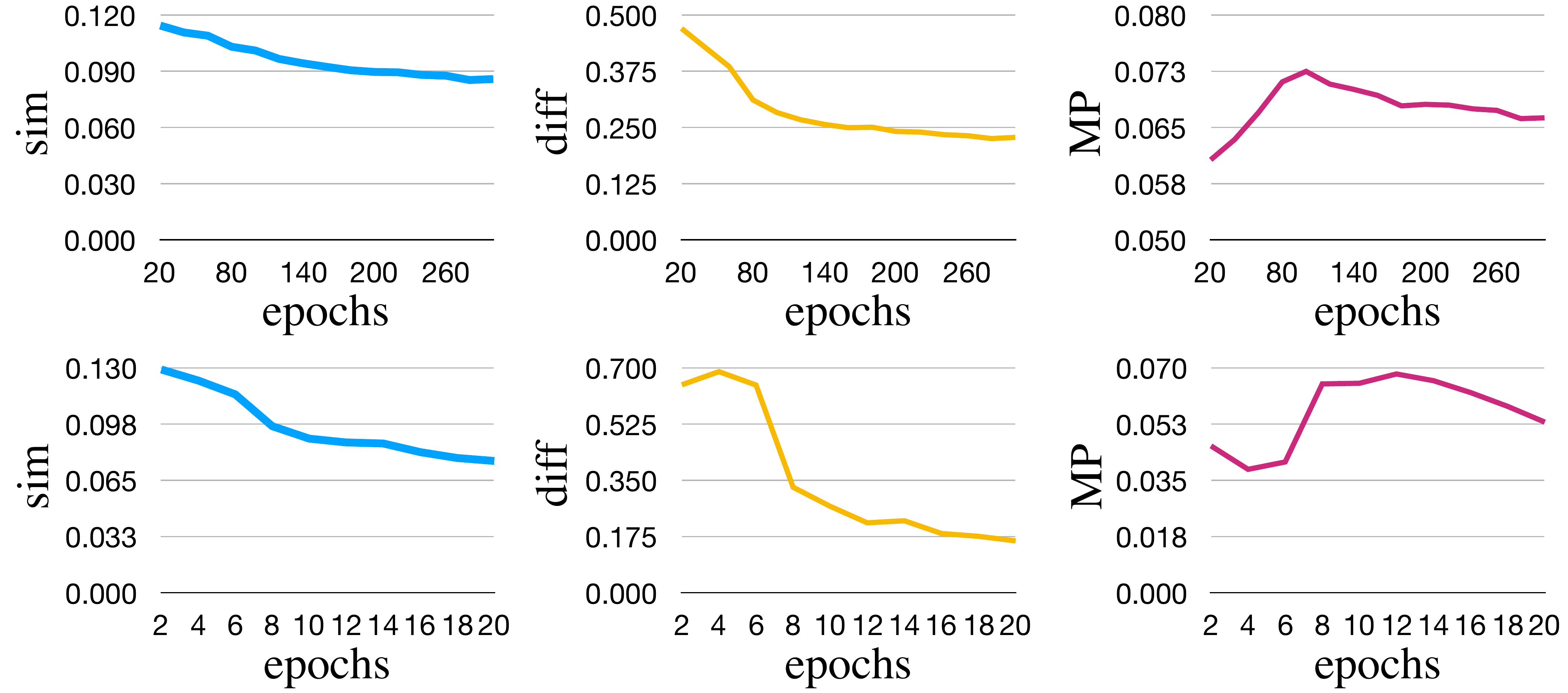}
\end{minipage}
\caption{Text-image similarity (sim), $L_{1}$ pixel difference (diff), and manipulative precision (MP) values at different epochs on the CUB (top) and COCO (bottom) datasets. We suggest to stop training the DCM module when the model gets the highest MP values shown in the last column.}
\label{fig:epochs}
\end{figure}

The main module is trained for 600 epochs on the CUB and 120 epochs on the COCO using the Adam optimiser \cite{kingma2014adam} with the learning rate 0.0002, and $\beta_{1}=0.5$, $\beta_{2}=0.999$. As for the detail correction module, there is a trade-off between the generation of new attributes corresponding to the given text and the reconstruction of text-irrelevant contents of the original image. Based on the manipulative precision (MP) values (see Fig.~\ref{fig:epochs}), we find that training 100 epochs for CUB, and 12 epochs for COCO to achieve an appropriate balance between generation and reconstruction. The other training setting is the same as in the main module. The hyperparameter controlling $\mathcal{L}_\text{reg}$ in Eq.~\eqref{equ:reg} is set to 1 for CUB and 15 for COCO. 

\smallskip\noindent\textbf{Manipulative precision metric.}
Image manipulation using natural language descriptions should be evaluated in terms of both the generation of new visual attributes from the given text, and the reconstruction of original contents existing in the input image. However, existing metrics only focus on one aspect of this problem. For example, the $L_{1}$ Euclidean distance, Peak Signal-to-Noise Ratio (PSNR), and SSIM \cite{wang2004image} only measure the similarity between two images, while the cosine similarity and the retrieval accuracy \cite{li2019controllable, nam2018text, xu2018attngan} only evaluate the similarity between the text and the corresponding generated image. Based on this, we contribute a new metric, called manipulative precision (MP), for this area to simultaneously measure the quality of generation and reconstruction. The metric is defined as
\begin{equation}
\text{MP} = (1-\text{diff}) \times \text{sim},
\label{equ:mp}
\end{equation}
where \text{diff} is the $L_{1}$ pixel difference between the input image and the corresponding modified image, \text{sim} is the text-image similarity, which is calculated by using pretrained text and image encoders \cite{xu2018attngan} based on a text-image matching score to extract global feature vectors of a given text description and the corresponding modified image, and then the similarity value is computed by applying cosine similarity between these two global vectors. Specifically, the design is based on the intuition that if the manipulated image is generated from an identity mapping network, then the text-image similarity should be low, as the synthetic image cannot perfectly keep a semantic consistency with the given text description.

\noindent\begin{figure*}[t]
\begin{minipage}{0.092\textwidth}
\raggedright
\scriptsize{Text}
\end{minipage}
\begin{minipage}{0.215\textwidth}
\centering
\scriptsize{This bird is \textbf{yellow} with a \textbf{yellow belly}, and has a \textbf{yellow beak}.}
\end{minipage}
\;\begin{minipage}{0.215\textwidth}
\centering
\scriptsize{A small bird with a \textbf{red belly}, a \textbf{red crown}, and \textbf{black wings}.}
\end{minipage}
\;\begin{minipage}{0.105\textwidth}
\centering
\scriptsize{Zebra, \textbf{dirt}.}
\end{minipage}
\begin{minipage}{0.105\textwidth}
\centering
\scriptsize{\textbf{Yellow} bus.}
\end{minipage}
\begin{minipage}{0.105\textwidth}
\centering
\scriptsize{\textbf{Evening}.}
\end{minipage}
\begin{minipage}{0.105\textwidth}
\centering
\scriptsize{Sheep, \textbf{dry grass}.}
\end{minipage}

\noindent\begin{minipage}{0.092\textwidth}
\raggedright
\scriptsize{Original}
\end{minipage}
\noindent\begin{minipage}{0.105\textwidth}
\includegraphics[width=1\linewidth, height=1\linewidth]{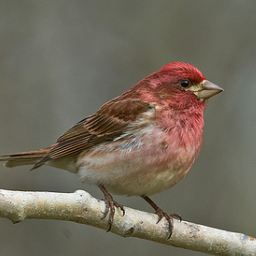}
\end{minipage}
\noindent\begin{minipage}{0.105\textwidth}
\includegraphics[width=1\linewidth, height=1\linewidth]{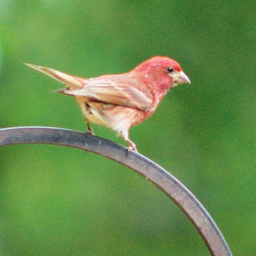}
\end{minipage}
\;\begin{minipage}{0.105\textwidth}
\includegraphics[width=1\linewidth, height=1\linewidth]{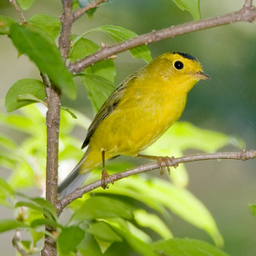}
\end{minipage}
\noindent\begin{minipage}{0.105\textwidth}
\includegraphics[width=1\linewidth, height=1\linewidth]{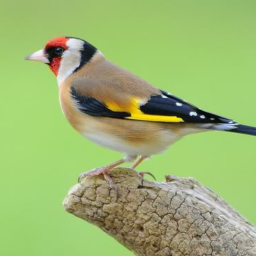}
\end{minipage}
\;\begin{minipage}{0.105\textwidth}
\includegraphics[width=1\linewidth, height=1\linewidth]{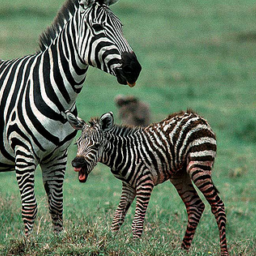}
\end{minipage}
\noindent\begin{minipage}{0.105\textwidth}
\includegraphics[width=1\linewidth, height=1\linewidth]{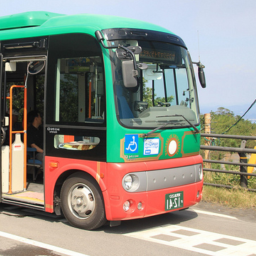}
\end{minipage}
\begin{minipage}{0.105\textwidth}
\includegraphics[width=1\linewidth, height=1\linewidth]{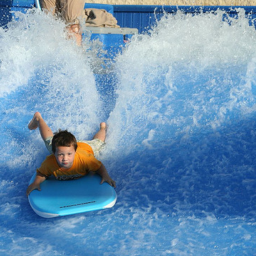}
\end{minipage}
\noindent\begin{minipage}{0.105\textwidth}
\includegraphics[width=1\linewidth, height=1\linewidth]{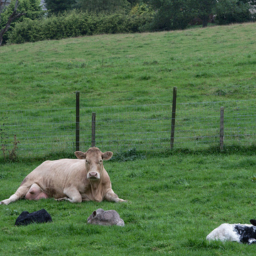}
\end{minipage}

\noindent\begin{minipage}{0.092\textwidth}
\raggedright
\scriptsize{SISGAN \cite{dong2017semantic}}
\end{minipage}
\noindent\begin{minipage}{0.105\textwidth}
\includegraphics[width=1\linewidth, height=1\linewidth]{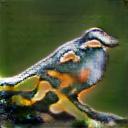}
\end{minipage}
\noindent\begin{minipage}{0.105\textwidth}
\includegraphics[width=1\linewidth, height=1\linewidth]{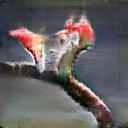}
\end{minipage}
\;\begin{minipage}{0.105\textwidth}
\includegraphics[width=1\linewidth, height=1\linewidth]{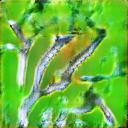}
\end{minipage}
\noindent\begin{minipage}{0.105\textwidth}
\includegraphics[width=1\linewidth, height=1\linewidth]{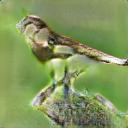}
\end{minipage}
\;\begin{minipage}{0.105\textwidth}
\includegraphics[width=1\linewidth, height=1\linewidth]{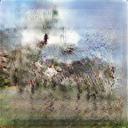}
\end{minipage}
\noindent\begin{minipage}{0.105\textwidth}
\includegraphics[width=1\linewidth, height=1\linewidth]{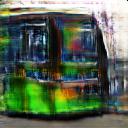}
\end{minipage}
\begin{minipage}{0.105\textwidth}
\includegraphics[width=1\linewidth, height=1\linewidth]{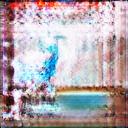}
\end{minipage}
\noindent\begin{minipage}{0.105\textwidth}
\includegraphics[width=1\linewidth, height=1\linewidth]{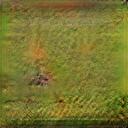}
\end{minipage}

\noindent\begin{minipage}{0.092\textwidth}
\raggedright
\scriptsize{TAGAN \cite{nam2018text}}
\end{minipage}
\noindent\begin{minipage}{0.105\textwidth}
\includegraphics[width=1\linewidth, height=1\linewidth]{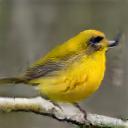}
\end{minipage}
\noindent\begin{minipage}{0.105\textwidth}
\includegraphics[width=1\linewidth, height=1\linewidth]{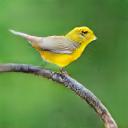}
\end{minipage}
\;\begin{minipage}{0.105\textwidth}
\includegraphics[width=1\linewidth, height=1\linewidth]{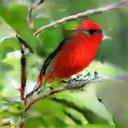}
\end{minipage}
\noindent\begin{minipage}{0.105\textwidth}
\includegraphics[width=1\linewidth, height=1\linewidth]{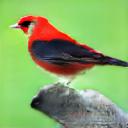}
\end{minipage}
\;\begin{minipage}{0.105\textwidth}
\includegraphics[width=1\linewidth, height=1\linewidth]{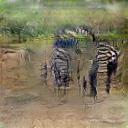}
\end{minipage}
\noindent\begin{minipage}{0.105\textwidth}
\includegraphics[width=1\linewidth, height=1\linewidth]{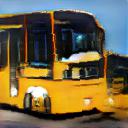}
\end{minipage}
\begin{minipage}{0.105\textwidth}
\includegraphics[width=1\linewidth, height=1\linewidth]{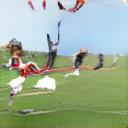}
\end{minipage}
\noindent\begin{minipage}{0.105\textwidth}
\includegraphics[width=1\linewidth, height=1\linewidth]{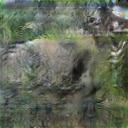}
\end{minipage}

\begin{minipage}{0.092\textwidth}
\raggedright
\scriptsize{Ours}
\end{minipage}
\begin{minipage}{0.105\textwidth}
\includegraphics[width=1\linewidth, height=1\linewidth]{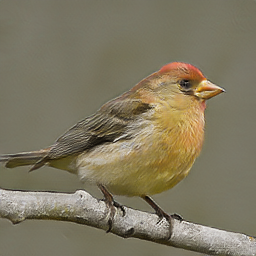}
\end{minipage}
\begin{minipage}{0.105\textwidth}
\includegraphics[width=1\linewidth, height=1\linewidth]{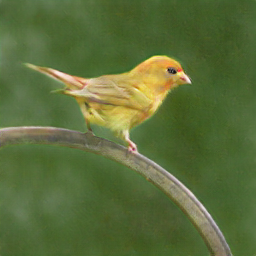}
\end{minipage}
\;\begin{minipage}{0.105\textwidth}
\includegraphics[width=1\linewidth, height=1\linewidth]{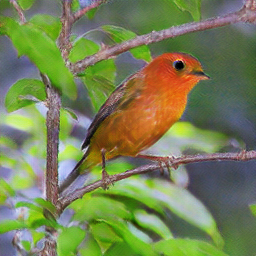}
\end{minipage}
\begin{minipage}{0.105\textwidth}
\includegraphics[width=1\linewidth, height=1\linewidth]{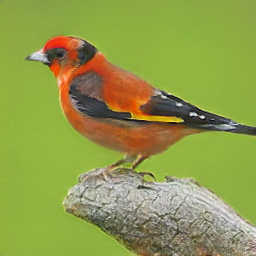}
\end{minipage}
\;\begin{minipage}{0.105\textwidth}
\includegraphics[width=1\linewidth, height=1\linewidth]{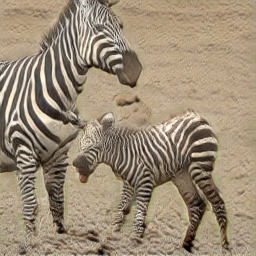}
\end{minipage}
\begin{minipage}{0.105\textwidth}
\includegraphics[width=1\linewidth, height=1\linewidth]{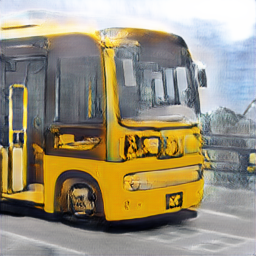}
\end{minipage}
\begin{minipage}{0.105\textwidth}
\includegraphics[width=1\linewidth, height=1\linewidth]{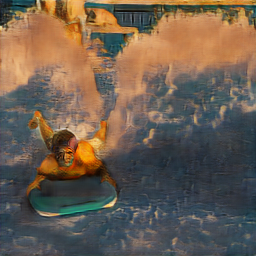}
\end{minipage}
\begin{minipage}{0.105\textwidth}
\includegraphics[width=1\linewidth, height=1\linewidth]{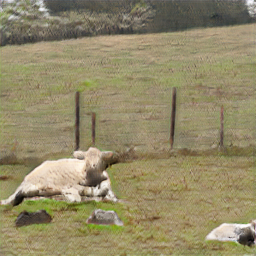}
\end{minipage}

\centering
\caption{Qualitative comparison of three methods on the CUB bird and COCO datasets.}
\label{fig:qual}
\end{figure*}

\vspace{-4mm}
\subsection{Comparison with state-of-the-art approaches}
\noindent\textbf{Quantitative comparison.}\label{sec:ex_quan}
As mentioned above, our model can generate high-quality images compared with the state-of-the-art methods. To demonstrate this, we adopt the inceptions score (IS) \cite{salimans2016improved} as a quantitative evaluation measure. Besides, we adopt manipulative precision (MP) to evaluate manipulation results.
In our experiments, we evaluate the IS on a large number of manipulated samples generated from mismatched pairs, i.e., randomly chosen input images manipulated by randomly selected text descriptions. 


As shown in Table~\ref{table:quantitative}, our method has the highest IS and MP values on both the CUB and COCO datasets compared with the state-of-the-art approaches, which demonstrates that (1) our method can produce high-quality manipulated results, and (2) our method can better generate new attributes matching the given text, and also effectively reconstruct text-irrelevant contents of the original image. 

\smallskip\noindent\textbf{Qualitative comparison.} Fig.~\ref{fig:qual} shows the visual comparison between our ManiGAN, SISGAN \cite{dong2017semantic}, and TAGAN \cite{nam2018text} on the CUB and COCO datasets. It can be seen that both state-of-the-art methods are only able to produce low-quality results and cannot effectively manipulate input images on the COCO dataset. However, our method is capable of performing an accurate manipulation and also keep a highly semantic consistency between synthetic images and given text descriptions, while preserving text-irrelevant contents. For example, shown in the last column of Fig.~\ref{fig:qual}, SISGAN and TAGAN both fail to achieve an effective manipulation, while our model modifies the \emph{green grass} to \emph{dry grass} and also edits the \emph{cow} into a \emph{sheep}. 

Note that as birds can have detailed descriptions (e.g., colour for different parts), we use a long sentence to manipulate them, while the descriptions for COCO are more abstract and focus mainly on categories, thus we use words (i.e., object + desired attributes) to do manipulation for simplicity, which has the same effect as using a sentence.
\begin{figure*}[t]
\begin{minipage}{0.105\textwidth}
\centering
\scriptsize{This bird has a \textbf{light grey belly}, \textbf{dark grey wings} and \textbf{head} with a \textbf{red beak}.}
\end{minipage}
\begin{minipage}{0.140\textwidth}
\includegraphics[width=1\linewidth, height=1\linewidth]{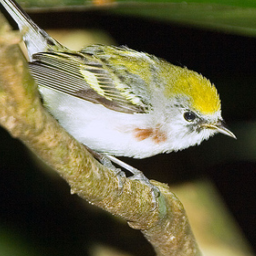}
\end{minipage}
\begin{minipage}{0.140\textwidth}
\includegraphics[width=1\linewidth, height=1\linewidth]{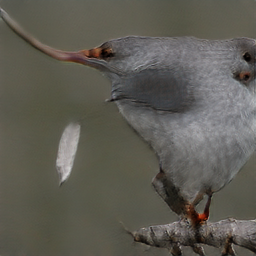}
\end{minipage}
\begin{minipage}{0.140\textwidth}
\includegraphics[width=1\linewidth, height=1\linewidth]{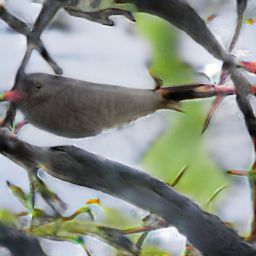}
\end{minipage}
\begin{minipage}{0.140\textwidth}
\includegraphics[width=1\linewidth, height=1\linewidth]{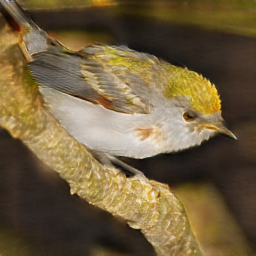}
\end{minipage}
\begin{minipage}{0.140\textwidth}
\includegraphics[width=1\linewidth, height=1\linewidth]{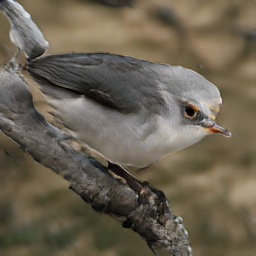}
\end{minipage}
\begin{minipage}{0.140\textwidth}
\includegraphics[width=1\linewidth, height=1\linewidth]{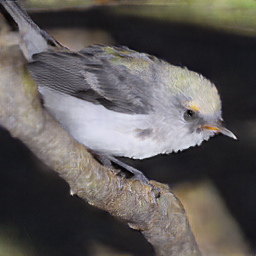}
\end{minipage}

\begin{minipage}{0.105\textwidth}
\centering
\scriptsize{This bird has a \textbf{yellow crown}, \textbf{blue wings} and a \textbf{yellow belly}.}
\end{minipage}
\begin{minipage}{0.140\textwidth}
\includegraphics[width=1\linewidth, height=1\linewidth]{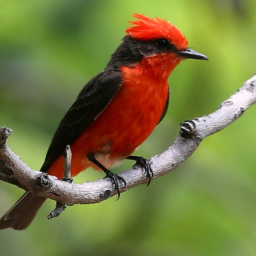}
\end{minipage}
\begin{minipage}{0.140\textwidth}
\includegraphics[width=1\linewidth, height=1\linewidth]{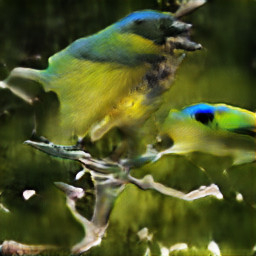}
\end{minipage}
\begin{minipage}{0.140\textwidth}
\includegraphics[width=1\linewidth, height=1\linewidth]{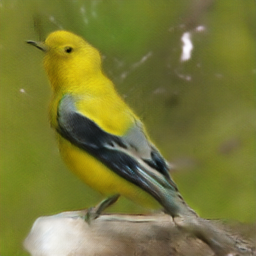}
\end{minipage}
\begin{minipage}{0.140\textwidth}
\includegraphics[width=1\linewidth, height=1\linewidth]{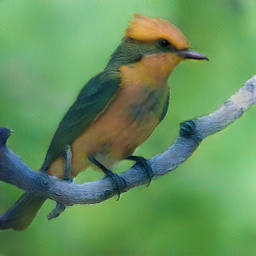}
\end{minipage}
\begin{minipage}{0.140\textwidth}
\includegraphics[width=1\linewidth, height=1\linewidth]{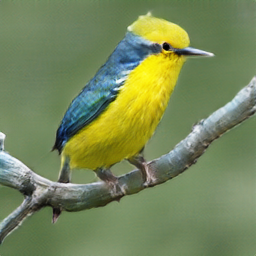}
\end{minipage}
\begin{minipage}{0.140\textwidth}
\includegraphics[width=1\linewidth, height=1\linewidth]{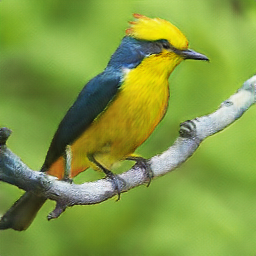}
\end{minipage}

\begin{minipage}{0.105\textwidth}
\centering
\scriptsize{Zebra, \textbf{green grass}.}
\end{minipage}
\begin{minipage}{0.140\textwidth}
\includegraphics[width=1\linewidth, height=1\linewidth]{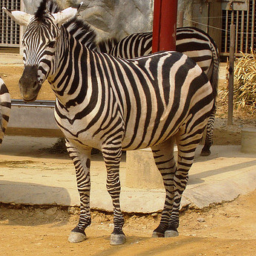}
\end{minipage}
\begin{minipage}{0.140\textwidth}
\includegraphics[width=1\linewidth, height=1\linewidth]{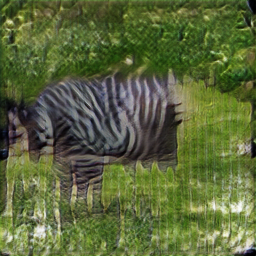}
\end{minipage}
\begin{minipage}{0.140\textwidth}
\includegraphics[width=1\linewidth, height=1\linewidth]{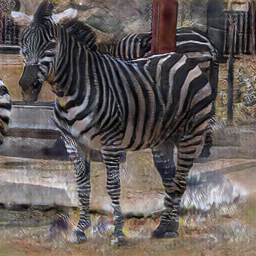}
\end{minipage}
\begin{minipage}{0.140\textwidth}
\includegraphics[width=1\linewidth, height=1\linewidth]{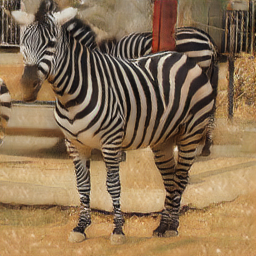}
\end{minipage}
\begin{minipage}{0.140\textwidth}
\includegraphics[width=1\linewidth, height=1\linewidth]{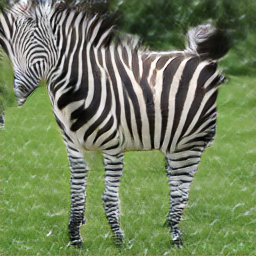}
\end{minipage}
\begin{minipage}{0.140\textwidth}
\includegraphics[width=1\linewidth, height=1\linewidth]{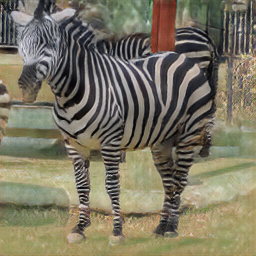}
\end{minipage}

\begin{minipage}{0.105\textwidth}
\centering
\scriptsize{\textbf{Yellow}, \textbf{green}, bus.}
\end{minipage}
\begin{minipage}{0.140\textwidth}
\includegraphics[width=1\linewidth, height=1\linewidth]{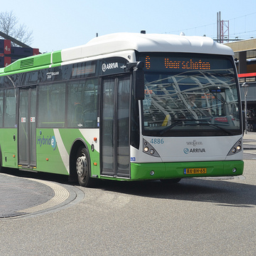}
\end{minipage}
\begin{minipage}{0.140\textwidth}
\includegraphics[width=1\linewidth, height=1\linewidth]{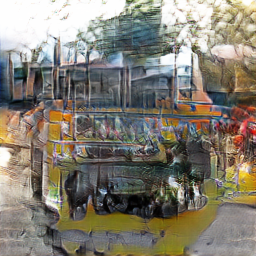}
\end{minipage}
\begin{minipage}{0.140\textwidth}
\includegraphics[width=1\linewidth, height=1\linewidth]{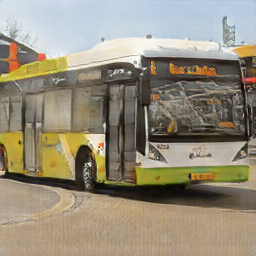}
\end{minipage}
\begin{minipage}{0.140\textwidth}
\includegraphics[width=1\linewidth, height=1\linewidth]{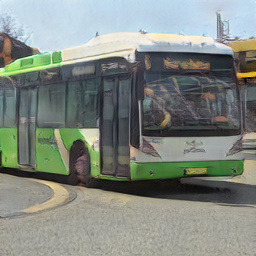}
\end{minipage}
\begin{minipage}{0.140\textwidth}
\includegraphics[width=1\linewidth, height=1\linewidth]{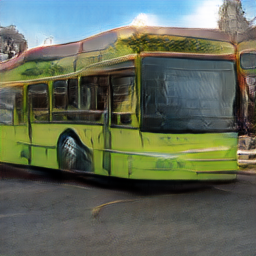}
\end{minipage}
\begin{minipage}{0.140\textwidth}
\includegraphics[width=1\linewidth, height=1\linewidth]{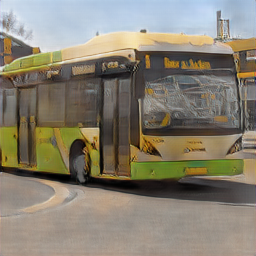}
\end{minipage}
\smallskip

\begin{minipage}{0.105\textwidth}
\centering
\scriptsize{a: Text}
\end{minipage}
\begin{minipage}{0.140\textwidth}
\centering
\scriptsize{b: Original}
\end{minipage}
\begin{minipage}{0.140\textwidth}
\centering
\scriptsize{c: Ours w/o ACM}
\end{minipage}
\begin{minipage}{0.140\textwidth}
\centering
\scriptsize{d: Our w/ Concat.}
\end{minipage}
\begin{minipage}{0.140\textwidth}
\centering
\scriptsize{e: Ours w/o main module}
\end{minipage}
\begin{minipage}{0.140\textwidth}
\centering
\scriptsize{f: Ours w/o DCM}
\end{minipage}
\begin{minipage}{0.140\textwidth}
\centering
\scriptsize{g: Ours}
\end{minipage}

\centering
\caption{Ablation studies. a: given text describing the desired visual attributes; b: input image; c: removing all ACMs and DCM, only concatenating image and text features before feeding into the main module; d: using the concatenation method to replace all ACMs; e: removing the main module and just training DCM only; f: removing DCM and just training the main module only; g: our full model.}
\label{fig:ablation}
\end{figure*}

\subsection{Ablation studies}
\label{sec:ablation}
\noindent\textbf{Ablation experiments of the affine combination module.}
To better understand what has been learned by our ACM, we ablate and visualise the learned feature maps shown in Fig.~\ref{fig:ana_vis}. As we can see, without $W$, some attributes cannot be perfectly generated (e.g., white belly in the first row and red head in the second row), and without $b$, the text-irrelevant contents (e.g., background) are hard to preserve, which verifies our assumption that $W$ behaves as a regional selection function to help the model focus on attributes corresponding to the given text, and $b$ helps to complete missing text-irrelevant details of the original image. Also, the visualisation of the channel feature maps of $W(v)$, $h \odot W(v)$, and $b(v)$ shown in the last three columns of Fig.~\ref{fig:ana_vis} validates the regional selection effect of the multiplication operation.

\smallskip\noindent\textbf{Effectiveness of the affine combination module.}
To verify the effectiveness of the ACM, we use the concatenation method to replace all ACMs, which concatenates hidden features $h$ and regional features $v$ along the channel direction, shown in Fig.~\ref{fig:ablation}~(d). As we can see, with the concatenation method, the model generates structurally different birds on CUB, and fails to do manipulation on COCO, which indicates that it is hard for the concatenation method to achieve a good balance between generation and reconstruction. The results on CUB is an example of the generation effect surpassing the reconstruction effect, while results on COCO show the domination of the reconstruction effect. In contrast, due to the regional selection effect of ACM that can distinguish which parts need to be generated or to be reconstructed, our full model synthesises an object having the same shape, pose, and position as the one existing in the original image, and also generates new visual attributes aligned with the given text description. 



Also, to further validate the effectiveness of ACM, we conduct an ablation study shown in Fig.~\ref{fig:ablation}~(c). In ``Our w/o ACM", we fully remove ACM in the main module and remove DCM as well. 
That is the main module without ACM, and we only concatenate original image features with text features at the beginning of the model and do not further provide additional original image features in the middle of the model. This method is used in both state-of-the-art SISGAN \cite{dong2017semantic} and TAGAN \cite{nam2018text}. It can be seen that our model without ACM fails to produce realistic images on both datasets. In contrast, our full model better generates attributes matching the given text, and also reconstructs text-irrelevant contents shown in (g). Table~\ref{table:quantitative} also verifies the effectiveness of our ACM, as the values of IS and MP increase significantly when we implement ACM.

\smallskip\noindent\textbf{Effectiveness of the detail correction module and main module.}\label{sec:component_dcm}
As shown in Fig.~\ref{fig:ablation}~(f), our model without DCM misses some attributes (e.g., the bird missing the tail in the second row, the zebra missing the mouth in the third row), or generates new contents (e.g., new background in the first row, different appearance of the bus in the fourth row), which indicates that our DCM can correct inappropriate attributes and reconstruct text-irrelevant contents. Fig.~\ref{fig:ablation}~(e) shows that without main module, our model fails to do image manipulation on both datasets, which just achieves an identity mapping. This is mainly because the model fails to correlate words with corresponding attributes, which has been done in the main module. Table~\ref{table:quantitative} also illustrates the identity mapping, as our model without main module gets the lowest $L_{1}$ pixel difference value.

\vspace{-1mm}
\section{Conclusion}
We have proposed a novel generative adversarial network for image manipulation, called ManiGAN, which can semantically manipulate input images using natural language descriptions. Two novel components are proposed: (1) the affine combination module selects image regions according to the given text, and then correlates the regions with corresponding semantic words for effective manipulation. Meanwhile, it encodes original image features for text-irrelevant contents reconstruction. (2) The detail correction module rectifies mismatched visual attributes and completes missing contents in the synthetic image. Extensive experimental results demonstrate the superiority of our method, in terms of both the effectiveness of image manipulation and the capability of generating high-quality results. 
{\small
\bibliographystyle{ieee_fullname}
\bibliography{egbib}
}
\newpage
\newpage
\appendix
\section{Architecture}

\begin{figure*}[t]
\centering
\begin{minipage}{1\textwidth}
\includegraphics[width=1\linewidth, height=0.475\linewidth]{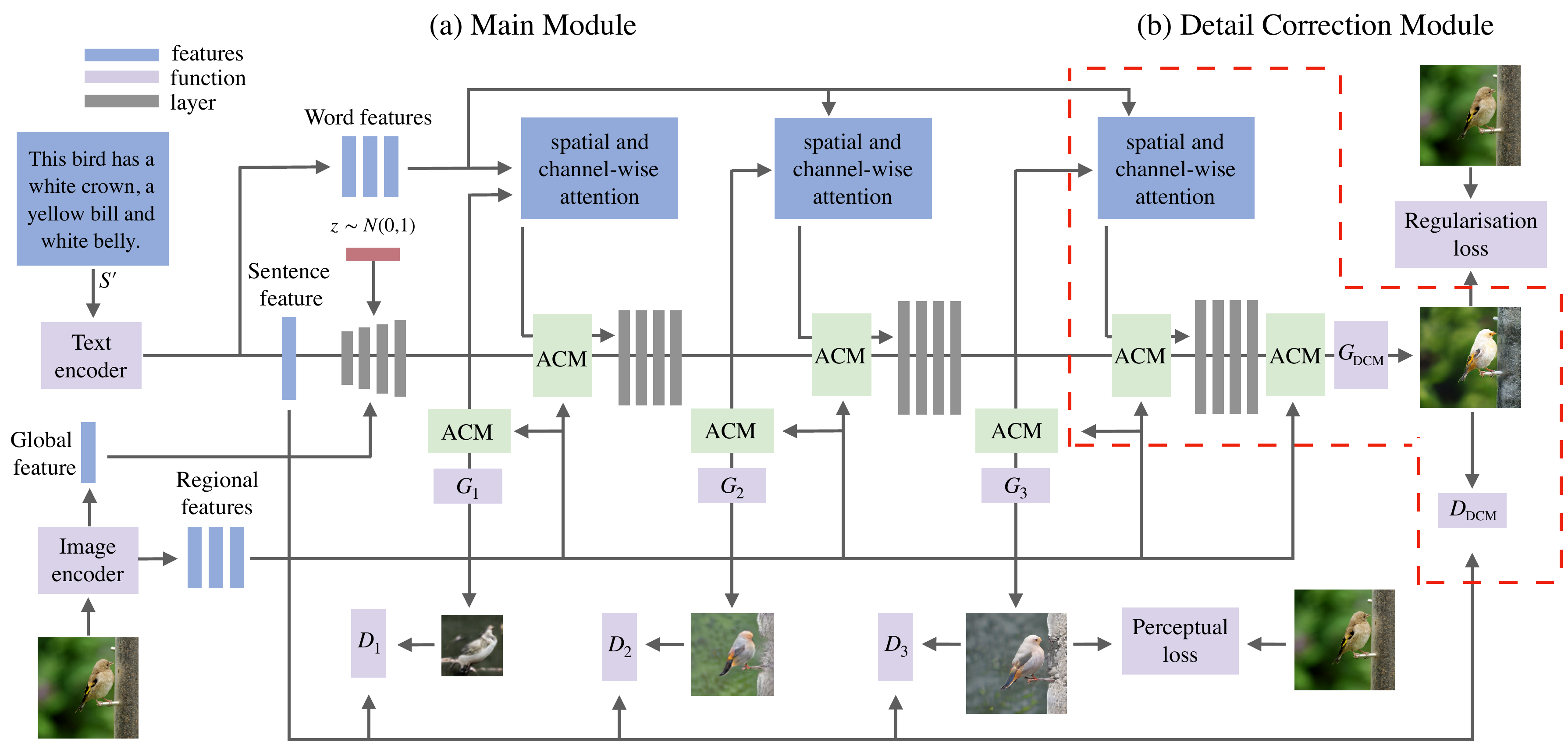}
\end{minipage}
\smallskip

\caption{The architecture of ManiGAN. ACM denotes the text-image affine combination module. Red dashed box indicates the architecture of the detail correction module.}
\label{fig:archi_full}
\end{figure*}

We adopt the ControlGAN \cite{li2019controllable} as the basic framework and replace batch normalisation with instance normalisation \cite{ulyanov2016instance} everywhere in the generator network except in the first stage. Basically, the affine combination module (ACM) can be inserted anywhere in the generator, but we experimentally find that it is best to incorporate the module before upsampling blocks and image generation networks; see Fig.~\ref{fig:archi_full}. 

\subsection{Residual Block}
\begin{figure}[h!]
\centering
\begin{minipage}{0.2\textwidth}
\includegraphics[width=0.8\linewidth, height=1.42\linewidth]{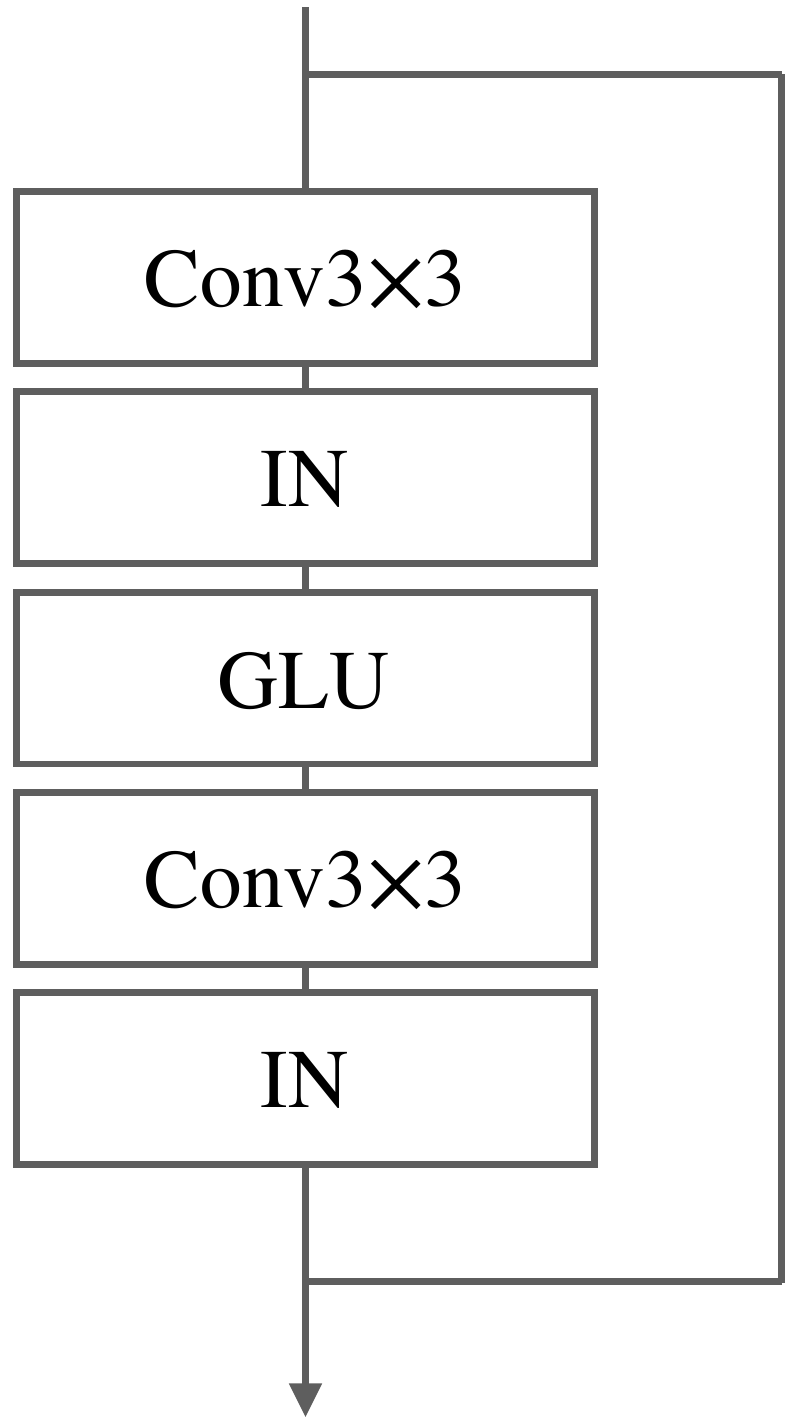}
\end{minipage}
\caption{The architecture of the residual block.}
\label{fig:archi_residual}
\end{figure}

Each residual block contains two convolutional layers, two instance normalisation (IN) \cite{ulyanov2016instance}, and one GLU \cite{dauphin2017language} non-linear function. The architecture of the residual block used in the detail correction module is shown in Fig.~\ref{fig:archi_residual}.

\bigskip

\begin{figure*}[h]
\centering
\begin{minipage}{0.113\textwidth}
\centering
\scriptsize{This bird has a \textbf{yellow bill}, a \textbf{blue head}, \textbf{blue wings} and a \textbf{yellow belly}.}
\end{minipage}
\begin{minipage}{0.113\textwidth}
\includegraphics[width=1\linewidth, height=1\linewidth]{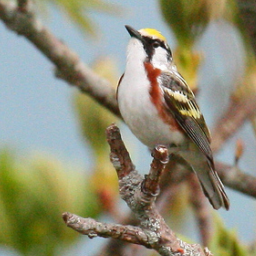}
\end{minipage}
\begin{minipage}{0.113\textwidth}
\includegraphics[width=1\linewidth, height=1\linewidth]{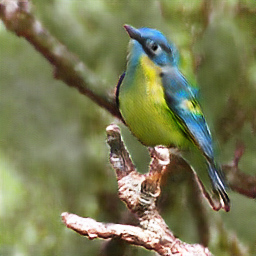}
\end{minipage}
\begin{minipage}{0.113\textwidth}
\includegraphics[width=1\linewidth, height=1\linewidth]{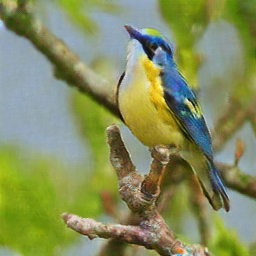}
\end{minipage}
\begin{minipage}{0.113\textwidth}
\includegraphics[width=1\linewidth, height=1\linewidth]{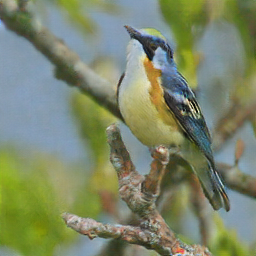}
\end{minipage}
\begin{minipage}{0.113\textwidth}
\includegraphics[width=1\linewidth, height=1\linewidth]{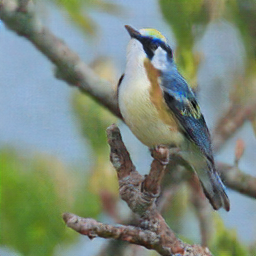}
\end{minipage}
\begin{minipage}{0.113\textwidth}
\includegraphics[width=1\linewidth, height=1\linewidth]{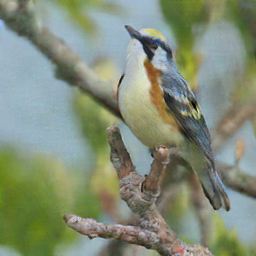}
\end{minipage}
\begin{minipage}{0.113\textwidth}
\includegraphics[width=1\linewidth, height=1\linewidth]{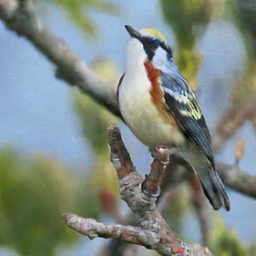}
\end{minipage}
\smallskip

\begin{minipage}{0.113\textwidth}
\centering
\scriptsize{Text}\end{minipage}
\begin{minipage}{0.113\textwidth}
\centering
\scriptsize{Original}\end{minipage}
\begin{minipage}{0.113\textwidth}
\centering
\scriptsize{50 epochs}
\end{minipage}
\begin{minipage}{0.113\textwidth}
\centering
\scriptsize{\textbf{100 epochs}}
\end{minipage}
\begin{minipage}{0.113\textwidth}
\centering
\scriptsize{150 epochs}
\end{minipage}
\begin{minipage}{0.113\textwidth}
\centering
\scriptsize{200 epochs}
\end{minipage}
\begin{minipage}{0.113\textwidth}
\centering
\scriptsize{250 epochs}
\end{minipage}
\begin{minipage}{0.113\textwidth}
\centering
\scriptsize{300 epochs}
\end{minipage}

\centering
\caption{Trend of the manipulation results over epoch increases on the CUB dataset.}
\label{fig:trend_bird}
\bigskip

\begin{minipage}{0.113\textwidth}
\centering
\scriptsize{Zebra, \textbf{dirt}.}
\end{minipage}
\begin{minipage}{0.113\textwidth}
\includegraphics[width=1\linewidth, height=1\linewidth]{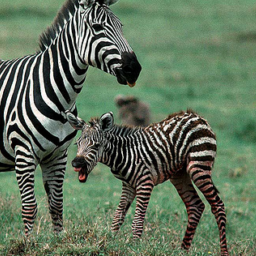}
\end{minipage}
\begin{minipage}{0.113\textwidth}
\includegraphics[width=1\linewidth, height=1\linewidth]{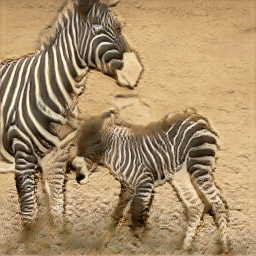}
\end{minipage}
\begin{minipage}{0.113\textwidth}
\includegraphics[width=1\linewidth, height=1\linewidth]{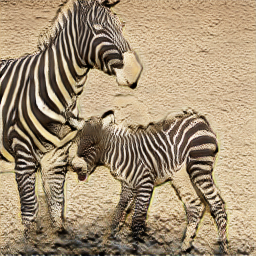}
\end{minipage}
\begin{minipage}{0.113\textwidth}
\includegraphics[width=1\linewidth, height=1\linewidth]{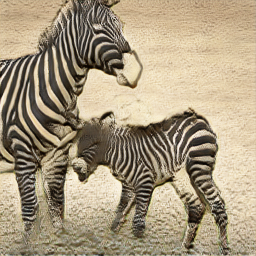}
\end{minipage}
\begin{minipage}{0.113\textwidth}
\includegraphics[width=1\linewidth, height=1\linewidth]{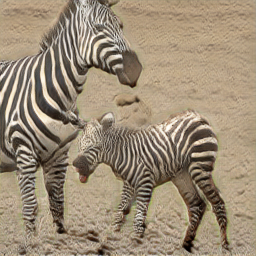}
\end{minipage}
\begin{minipage}{0.113\textwidth}
\includegraphics[width=1\linewidth, height=1\linewidth]{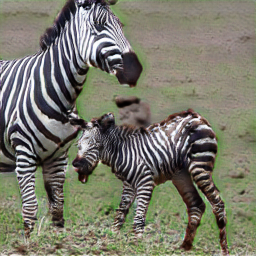}
\end{minipage}
\begin{minipage}{0.113\textwidth}
\includegraphics[width=1\linewidth, height=1\linewidth]{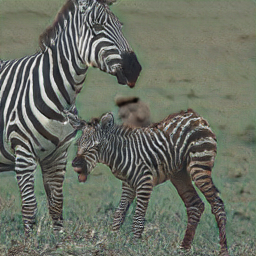}
\end{minipage}
\smallskip

\begin{minipage}{0.113\textwidth}
\centering
\scriptsize{Text}\end{minipage}
\begin{minipage}{0.113\textwidth}
\centering
\scriptsize{Original}\end{minipage}
\begin{minipage}{0.113\textwidth}
\centering
\scriptsize{3 epochs}
\end{minipage}
\begin{minipage}{0.113\textwidth}
\centering
\scriptsize{6 epochs}
\end{minipage}
\begin{minipage}{0.113\textwidth}
\centering
\scriptsize{9 epochs}
\end{minipage}
\begin{minipage}{0.113\textwidth}
\centering
\scriptsize{\textbf{12 epochs}}
\end{minipage}
\begin{minipage}{0.113\textwidth}
\centering
\scriptsize{15 epochs}
\end{minipage}
\begin{minipage}{0.113\textwidth}
\centering
\scriptsize{18 epochs}
\end{minipage}

\centering
\caption{Trend of the manipulation results over epoch increases on the COCO dataset.}
\label{fig:trend_coco}
\end{figure*}

\section{Objective Functions}
\label{apex:objective}
We train the main module and detail correction module separately, and the generator and discriminator in both modules are trained alternatively by minimising both the generator loss $\mathcal{L}_G$ and the discriminator loss $\mathcal{L}_D$.
\smallskip

\noindent\textbf{Generator objective.}
The loss function for the generator follows those used in ControlGAN \cite{li2019controllable}, but we introduce a regularisation term:
\begin{equation}
\mathcal{L}_\text{reg}=1-\frac{1}{CHW}|| {I}'-I ||
\textrm{,}
\end{equation}
to prevent the network achieving identity mapping, which can penalise large perturbations when the generated image becomes the same as the input image.
\begin{equation}
\footnotesize
\begin{split}
\mathcal{L}_{G}=&\underbrace{-\frac{1}{2}E_{{I}'\sim PG}\left [ \log(D({I}')) \right ]}_\text{unconditional adversarial loss}\underbrace{-\frac{1}{2}E_{{I}'\sim PG}\left [ \log(D({I}',S)) \right ]}_\text{conditional adversarial loss} \\
&+ \mathcal{L}_\text{ControlGAN} + \lambda_{1}\mathcal{L}_\text{reg}
\textrm{,}
\end{split}
\end{equation}
\begin{equation}
\footnotesize
\begin{split}
\mathcal{L}_\text{ControlGAN}=\lambda_{2}\mathcal{L}_\text{DAMSM} +\lambda_{3}(1-\mathcal{L}_\text{corre}(I',S))+\lambda_{4}\mathcal{L}_\text{rec}({I}', I) \textrm{,}
\end{split}
\label{equ:generator}
\end{equation}
where $I$ is the real image sampled from the true image distribution $P_{\text{data}}$, $S$ is the corresponding matched text that correctly describes the $I$, $I'$ is the generated image sampled from the model distribution $PG$. The unconditional adversarial loss makes the synthetic image $I'$ indistinguishable from the real image $I$, the conditional adversarial loss aligns the generated image $I'$ with the given text description $S$, $\mathcal{L}_\text{DAMSM}$ \cite{xu2018attngan} measures the text-image similarity at the word-level to provide fine-grained feedback for image generation, $\mathcal{L}_\text{corre}$ \cite{li2019controllable} determines whether word-related visual attributes exist in the image, and $\mathcal{L}_\text{rec}$ \cite{li2019controllable} reduces randomness involved in the generation process. $\lambda_{1}$, $\lambda_{2}$, $\lambda_{3}$, and $\lambda_{4}$ are hyperparameters controlling the importance of additional losses. Note that we do not use $\mathcal{L}_\text{rec}$ when we train the detail correction module.
\smallskip

\noindent\textbf{Discriminator objective.}
The loss function for the discriminator follows those used in ControlGAN \cite{li2019controllable}, and the function used to train the discriminator in the detail correction module is the same as the one used in the last stage of the main module.
\begin{equation}
\footnotesize
\begin{split}
\mathcal{L}_{D}=&\underbrace{-\frac{1}{2}E_{I\sim P_\text{data}}\left [ \log(D(I)) \right ]-\frac{1}{2}E_{{I}'\sim PG}\left [ \log(1-D({I}')) \right ]}_\text{unconditional adversarial loss}\\
&\underbrace{-\frac{1}{2}E_{I\sim P_\text{data}}\left [ \log(D(I,S)) \right ]-\frac{1}{2}E_{{I}'\sim PG}\left [ \log(1-D({I}',S)) \right ]}_\text{conditional adversarial loss}\\
&+\lambda_{3}((1-\mathcal{L}_\text{corre}(I,S))+ \mathcal{L}_\text{corre}(I,{S}'))
\textrm{,}
\end{split}
\end{equation}
where ${S}'$ is a given text description randomly sampled from the dataset. The unconditional adversarial loss determines whether the given image is real, and the conditional adversarial loss reflects the semantic similarity between images and texts.


\section{Trend of Manipulation Results}
We track the trend of manipulation results over epoch increases, as shown in Figs.~\ref{fig:trend_bird} and \ref{fig:trend_coco}. The original images are smoothly modified to achieve the best balance between the generation of new visual attributes (e.g., blue head, blue wings and yellow belly in Fig.~\ref{fig:trend_bird}, dirt background in Fig.~\ref{fig:trend_coco}) and the reconstruction of text-irrelevant contents of the original images (e.g., the shape of the bird and the background in Fig.~\ref{fig:trend_bird}, the appearance of zebras in Fig.~\ref{fig:trend_coco}). However, when the epoch goes larger, the generated visual attributes (e.g., blue head, blue wings, and yellow belly of the bird, dirt background of the zebras) aligned with the given text descriptions are gradually erased, and the synthetic images become more and more similar to the original images. This verifies the existence of the trade-off between the generation of new visual attributes required in the given text descriptions and the reconstruction of text-irrelevant contents existing in the original images.

\section{Additional Comparison Results}
\begin{figure*}[h!]
\begin{minipage}{1\textwidth}
\begin{minipage}{0.18\textwidth}
\centering
\small{This bird is \textbf{blue} and \textbf{grey} with a \textbf{red belly}.}
\end{minipage}
\;\;\begin{minipage}{0.195\textwidth}
\includegraphics[width=1\linewidth, height=1\linewidth]{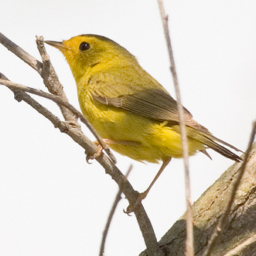}
\end{minipage}
\noindent\begin{minipage}{0.195\textwidth}
\includegraphics[width=1\linewidth, height=1\linewidth]{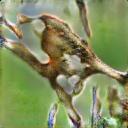}
\end{minipage}
\noindent\begin{minipage}{0.195\textwidth}
\includegraphics[width=1\linewidth, height=1\linewidth]{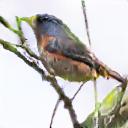}
\end{minipage}
\noindent\begin{minipage}{0.195\textwidth}
\includegraphics[width=1\linewidth, height=1\linewidth]{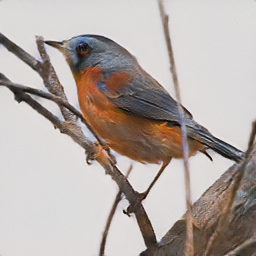}
\end{minipage}
\end{minipage}
\smallskip

\begin{minipage}{1\textwidth}
\begin{minipage}{0.18\textwidth}
\centering
\small{This bird has wings that are \textbf{grey} and \textbf{yellow} with a \textbf{yellow belly}.}
\end{minipage}
\;\;\begin{minipage}{0.195\textwidth}
\includegraphics[width=1\linewidth, height=1\linewidth]{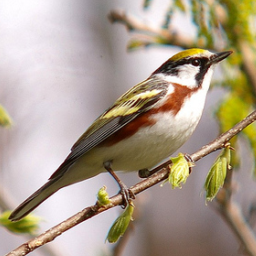}
\end{minipage}
\noindent\begin{minipage}{0.195\textwidth}
\includegraphics[width=1\linewidth, height=1\linewidth]{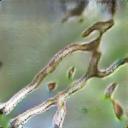}
\end{minipage}
\noindent\begin{minipage}{0.195\textwidth}
\includegraphics[width=1\linewidth, height=1\linewidth]{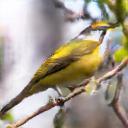}
\end{minipage}
\noindent\begin{minipage}{0.195\textwidth}
\includegraphics[width=1\linewidth, height=1\linewidth]{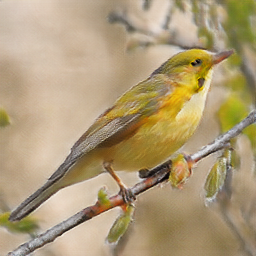}
\end{minipage}
\end{minipage}
\smallskip

\begin{minipage}{1\textwidth}
\begin{minipage}{0.18\textwidth}
\centering
\small{This bird is \textbf{black} in colour, with a \textbf{red crown} and a \textbf{red beak}.}
\end{minipage}
\;\;\begin{minipage}{0.195\textwidth}
\includegraphics[width=1\linewidth, height=1\linewidth]{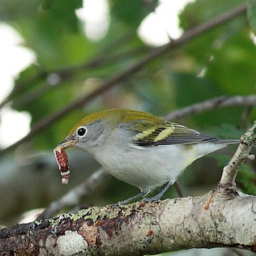}
\end{minipage}
\noindent\begin{minipage}{0.195\textwidth}
\includegraphics[width=1\linewidth, height=1\linewidth]{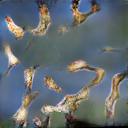}
\end{minipage}
\noindent\begin{minipage}{0.195\textwidth}
\includegraphics[width=1\linewidth, height=1\linewidth]{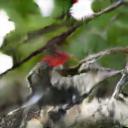}
\end{minipage}
\noindent\begin{minipage}{0.195\textwidth}
\includegraphics[width=1\linewidth, height=1\linewidth]{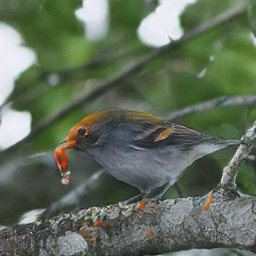}
\end{minipage}
\end{minipage}
\smallskip

\begin{minipage}{1\textwidth}
\begin{minipage}{0.18\textwidth}
\centering
\small{This green bird has \textbf{a black crown} and a \textbf{green belly}.}
\end{minipage}
\;\;\begin{minipage}{0.195\textwidth}
\includegraphics[width=1\linewidth, height=1\linewidth]{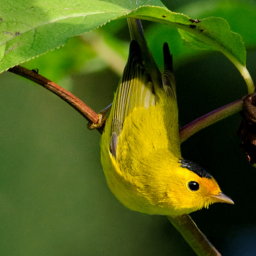}
\end{minipage}
\noindent\begin{minipage}{0.195\textwidth}
\includegraphics[width=1\linewidth, height=1\linewidth]{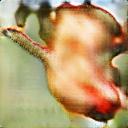}
\end{minipage}
\noindent\begin{minipage}{0.195\textwidth}
\includegraphics[width=1\linewidth, height=1\linewidth]{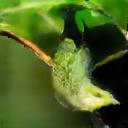}
\end{minipage}
\noindent\begin{minipage}{0.195\textwidth}
\includegraphics[width=1\linewidth, height=1\linewidth]{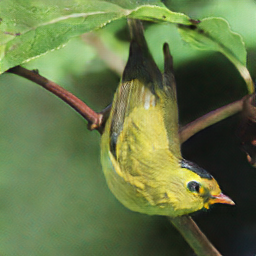}
\end{minipage}
\end{minipage}
\smallskip

\begin{minipage}{1\textwidth}
\begin{minipage}{0.18\textwidth}
\centering
\small{A bird with \textbf{brown wings} and a \textbf{yellow body}, with a \textbf{yellow head}.}
\end{minipage}
\;\;\begin{minipage}{0.195\textwidth}
\includegraphics[width=1\linewidth, height=1\linewidth]{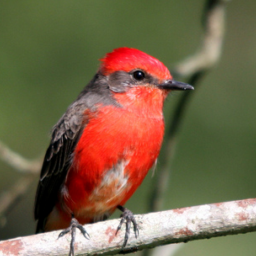}
\end{minipage}
\noindent\begin{minipage}{0.195\textwidth}
\includegraphics[width=1\linewidth, height=1\linewidth]{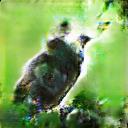}
\end{minipage}
\noindent\begin{minipage}{0.195\textwidth}
\includegraphics[width=1\linewidth, height=1\linewidth]{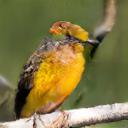}
\end{minipage}
\noindent\begin{minipage}{0.195\textwidth}
\includegraphics[width=1\linewidth, height=1\linewidth]{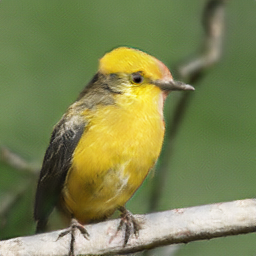}
\end{minipage}
\end{minipage}
\smallskip

\begin{minipage}{1\textwidth}
\begin{minipage}{0.18\textwidth}
\centering
\small{A white bird with \textbf{grey wings} and a \textbf{red bill}, with a \textbf{white belly}.}
\end{minipage}
\;\;\begin{minipage}{0.195\textwidth}
\includegraphics[width=1\linewidth, height=1\linewidth]{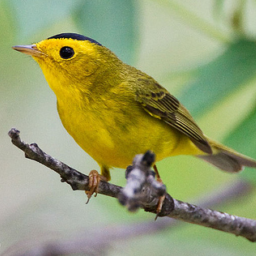}
\end{minipage}
\noindent\begin{minipage}{0.195\textwidth}
\includegraphics[width=1\linewidth, height=1\linewidth]{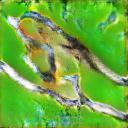}
\end{minipage}
\noindent\begin{minipage}{0.195\textwidth}
\includegraphics[width=1\linewidth, height=1\linewidth]{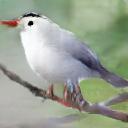}
\end{minipage}
\noindent\begin{minipage}{0.195\textwidth}
\includegraphics[width=1\linewidth, height=1\linewidth]{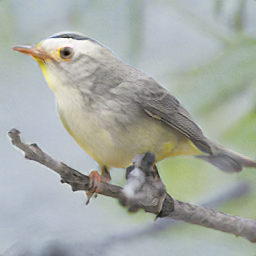}
\end{minipage}
\end{minipage}
\smallskip

\begin{minipage}{1\textwidth}
\begin{minipage}{0.195\textwidth}
\centering
\small{Given Text}
\end{minipage}
\begin{minipage}{0.195\textwidth}
\centering
\small{Original}
\end{minipage}
\noindent\begin{minipage}{0.195\textwidth}
\centering
\small{SISGAN \cite{dong2017semantic}}
\end{minipage}
\noindent\begin{minipage}{0.195\textwidth}
\centering
\small{TAGAN \cite{nam2018text}}
\end{minipage}
\noindent\begin{minipage}{0.195\textwidth}
\centering
\small{Ours}
\end{minipage}
\end{minipage}

\centering
\caption{Additional comparison results between ManiGAN, SISGAN, and TAGAN on the CUB bird dataset.}
\label{fig:qual_show1}
\end{figure*}

\begin{figure*}[h!]
\begin{minipage}{1\textwidth}
\begin{minipage}{0.18\textwidth}
\centering
\small{A small \textbf{blue} bird with an \textbf{orange crown}, with a \textbf{grey belly}.}
\end{minipage}
\;\;\begin{minipage}{0.195\textwidth}
\includegraphics[width=1\linewidth, height=1\linewidth]{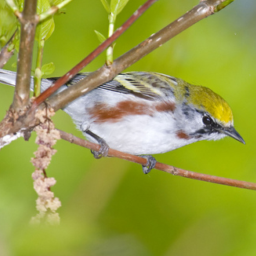}
\end{minipage}
\noindent\begin{minipage}{0.195\textwidth}
\includegraphics[width=1\linewidth, height=1\linewidth]{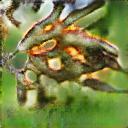}
\end{minipage}
\noindent\begin{minipage}{0.195\textwidth}
\includegraphics[width=1\linewidth, height=1\linewidth]{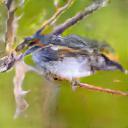}
\end{minipage}
\noindent\begin{minipage}{0.195\textwidth}
\includegraphics[width=1\linewidth, height=1\linewidth]{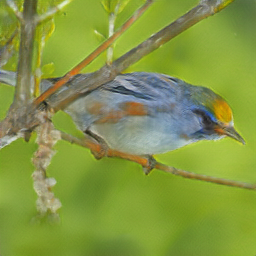}
\end{minipage}
\end{minipage}
\smallskip

\begin{minipage}{1\textwidth}
\begin{minipage}{0.18\textwidth}
\centering
\small{This bird has a \textbf{red head}, \textbf{black eye rings}, and a \textbf{yellow belly}.}
\end{minipage}
\;\;\begin{minipage}{0.195\textwidth}
\includegraphics[width=1\linewidth, height=1\linewidth]{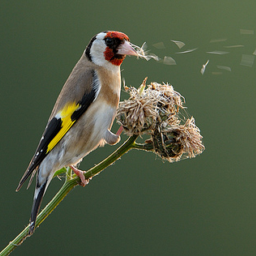}
\end{minipage}
\noindent\begin{minipage}{0.195\textwidth}
\includegraphics[width=1\linewidth, height=1\linewidth]{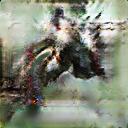}
\end{minipage}
\noindent\begin{minipage}{0.195\textwidth}
\includegraphics[width=1\linewidth, height=1\linewidth]{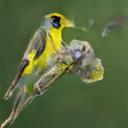}
\end{minipage}
\noindent\begin{minipage}{0.195\textwidth}
\includegraphics[width=1\linewidth, height=1\linewidth]{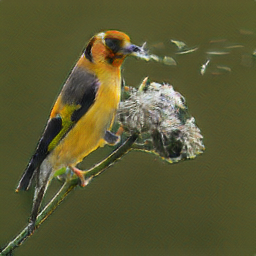}
\end{minipage}
\end{minipage}
\smallskip

\begin{minipage}{1\textwidth}
\begin{minipage}{0.18\textwidth}
\centering
\small{This bird is mostly \textbf{red} with a \textbf{black beak}, and a \textbf{black tail}.}
\end{minipage}
\;\;\begin{minipage}{0.195\textwidth}
\includegraphics[width=1\linewidth, height=1\linewidth]{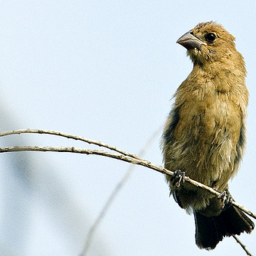}
\end{minipage}
\noindent\begin{minipage}{0.195\textwidth}
\includegraphics[width=1\linewidth, height=1\linewidth]{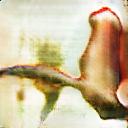}
\end{minipage}
\noindent\begin{minipage}{0.195\textwidth}
\includegraphics[width=1\linewidth, height=1\linewidth]{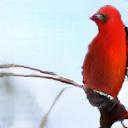}
\end{minipage}
\noindent\begin{minipage}{0.195\textwidth}
\includegraphics[width=1\linewidth, height=1\linewidth]{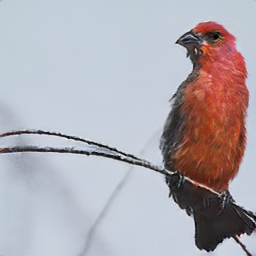}
\end{minipage}
\end{minipage}
\smallskip

\begin{minipage}{1\textwidth}
\begin{minipage}{0.18\textwidth}
\centering
\small{This tiny bird is \textbf{blue} and has a \textbf{red bill} and a \textbf{red belly}.}
\end{minipage}
\;\;\begin{minipage}{0.195\textwidth}
\includegraphics[width=1\linewidth, height=1\linewidth]{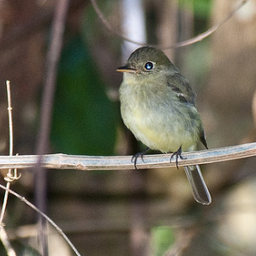}
\end{minipage}
\noindent\begin{minipage}{0.195\textwidth}
\includegraphics[width=1\linewidth, height=1\linewidth]{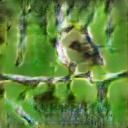}
\end{minipage}
\noindent\begin{minipage}{0.195\textwidth}
\includegraphics[width=1\linewidth, height=1\linewidth]{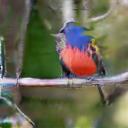}
\end{minipage}
\noindent\begin{minipage}{0.195\textwidth}
\includegraphics[width=1\linewidth, height=1\linewidth]{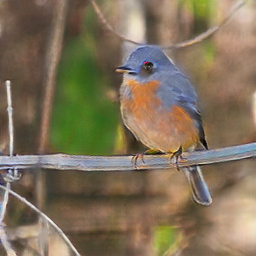}
\end{minipage}
\end{minipage}
\smallskip

\begin{minipage}{1\textwidth}
\begin{minipage}{0.18\textwidth}
\centering
\small{This bird has a \textbf{white head}, a \textbf{yellow bill}, and a \textbf{yellow belly}.}
\end{minipage}
\;\;\begin{minipage}{0.195\textwidth}
\includegraphics[width=1\linewidth, height=1\linewidth]{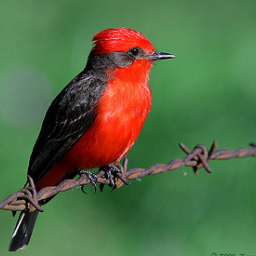}
\end{minipage}
\noindent\begin{minipage}{0.195\textwidth}
\includegraphics[width=1\linewidth, height=1\linewidth]{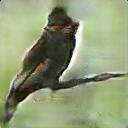}
\end{minipage}
\noindent\begin{minipage}{0.195\textwidth}
\includegraphics[width=1\linewidth, height=1\linewidth]{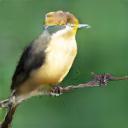}
\end{minipage}
\noindent\begin{minipage}{0.195\textwidth}
\includegraphics[width=1\linewidth, height=1\linewidth]{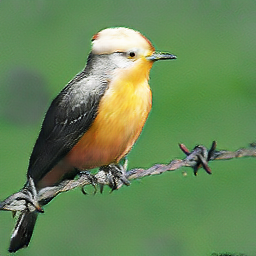}
\end{minipage}
\end{minipage}
\smallskip

\begin{minipage}{1\textwidth}
\begin{minipage}{0.18\textwidth}
\centering
\small{A white bird with \textbf{red throat}, \textbf{black eye rings}, and \textbf{grey wings}.}
\end{minipage}
\;\;\begin{minipage}{0.195\textwidth}
\includegraphics[width=1\linewidth, height=1\linewidth]{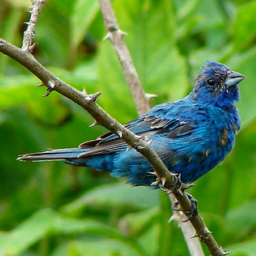}
\end{minipage}
\noindent\begin{minipage}{0.195\textwidth}
\includegraphics[width=1\linewidth, height=1\linewidth]{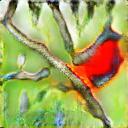}
\end{minipage}
\noindent\begin{minipage}{0.195\textwidth}
\includegraphics[width=1\linewidth, height=1\linewidth]{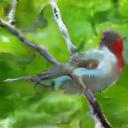}
\end{minipage}
\noindent\begin{minipage}{0.195\textwidth}
\includegraphics[width=1\linewidth, height=1\linewidth]{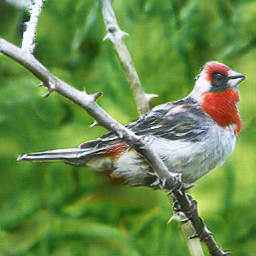}
\end{minipage}
\end{minipage}
\smallskip

\begin{minipage}{1\textwidth}
\begin{minipage}{0.195\textwidth}
\centering
\small{Given Text}
\end{minipage}
\begin{minipage}{0.195\textwidth}
\centering
\small{Original}
\end{minipage}
\noindent\begin{minipage}{0.195\textwidth}
\centering
\small{SISGAN \cite{dong2017semantic}}
\end{minipage}
\noindent\begin{minipage}{0.195\textwidth}
\centering
\small{TAGAN \cite{nam2018text}}
\end{minipage}
\noindent\begin{minipage}{0.195\textwidth}
\centering
\small{Ours}
\end{minipage}
\end{minipage}

\centering
\caption{Additional comparison results between ManiGAN, SISGAN, and TAGAN on the CUB bird dataset.}
\label{fig:qual_show2}
\end{figure*}

\begin{figure*}[h!]
\begin{minipage}{1\textwidth}
\begin{minipage}{0.18\textwidth}
\centering
\small{\textbf{Sunset}.}
\end{minipage}
\;\;\begin{minipage}{0.195\textwidth}
\includegraphics[width=1\linewidth, height=1\linewidth]{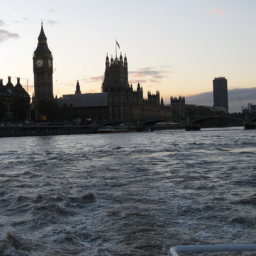}
\end{minipage}
\noindent\begin{minipage}{0.195\textwidth}
\includegraphics[width=1\linewidth, height=1\linewidth]{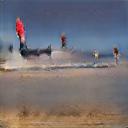}
\end{minipage}
\noindent\begin{minipage}{0.195\textwidth}
\includegraphics[width=1\linewidth, height=1\linewidth]{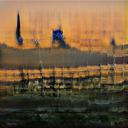}
\end{minipage}
\noindent\begin{minipage}{0.195\textwidth}
\includegraphics[width=1\linewidth, height=1\linewidth]{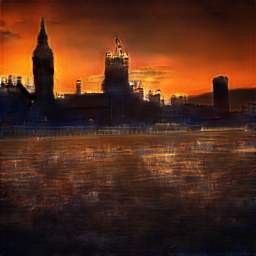}
\end{minipage}
\end{minipage}
\smallskip

\begin{minipage}{1\textwidth}
\begin{minipage}{0.18\textwidth}
\centering
\small{\textbf{Blue} boat, \textbf{green grass}.}
\end{minipage}
\;\;\begin{minipage}{0.195\textwidth}
\includegraphics[width=1\linewidth, height=1\linewidth]{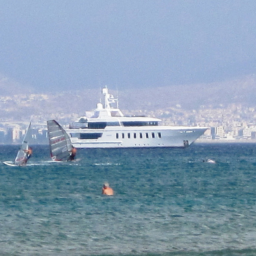}
\end{minipage}
\noindent\begin{minipage}{0.195\textwidth}
\includegraphics[width=1\linewidth, height=1\linewidth]{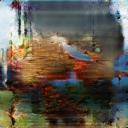}
\end{minipage}
\noindent\begin{minipage}{0.195\textwidth}
\includegraphics[width=1\linewidth, height=1\linewidth]{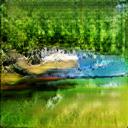}
\end{minipage}
\noindent\begin{minipage}{0.195\textwidth}
\includegraphics[width=1\linewidth, height=1\linewidth]{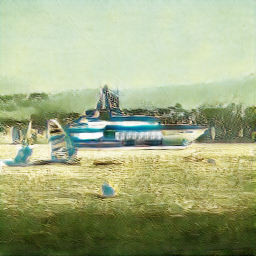}
\end{minipage}
\end{minipage}
\smallskip

\begin{minipage}{1\textwidth}
\begin{minipage}{0.18\textwidth}
\centering
\small{\textbf{White} bus.}
\end{minipage}
\;\;\begin{minipage}{0.195\textwidth}
\includegraphics[width=1\linewidth, height=1\linewidth]{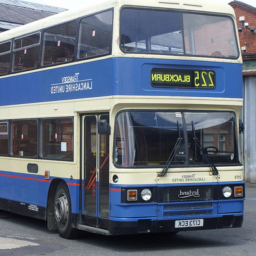}
\end{minipage}
\noindent\begin{minipage}{0.195\textwidth}
\includegraphics[width=1\linewidth, height=1\linewidth]{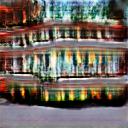}
\end{minipage}
\noindent\begin{minipage}{0.195\textwidth}
\includegraphics[width=1\linewidth, height=1\linewidth]{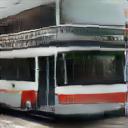}
\end{minipage}
\noindent\begin{minipage}{0.195\textwidth}
\includegraphics[width=1\linewidth, height=1\linewidth]{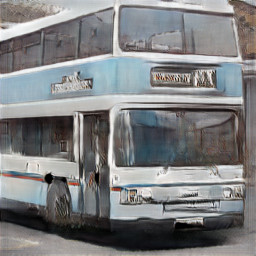}
\end{minipage}
\end{minipage}
\smallskip

\begin{minipage}{1\textwidth}
\begin{minipage}{0.18\textwidth}
\centering
\small{Man, \textbf{dry grass}.}
\end{minipage}
\;\;\begin{minipage}{0.195\textwidth}
\includegraphics[width=1\linewidth, height=1\linewidth]{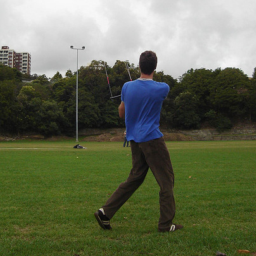}
\end{minipage}
\noindent\begin{minipage}{0.195\textwidth}
\includegraphics[width=1\linewidth, height=1\linewidth]{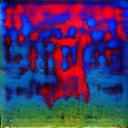}
\end{minipage}
\noindent\begin{minipage}{0.195\textwidth}
\includegraphics[width=1\linewidth, height=1\linewidth]{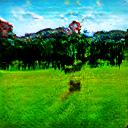}
\end{minipage}
\noindent\begin{minipage}{0.195\textwidth}
\includegraphics[width=1\linewidth, height=1\linewidth]{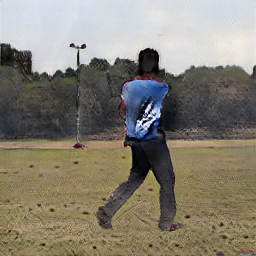}
\end{minipage}
\end{minipage}
\smallskip

\begin{minipage}{1\textwidth}
\begin{minipage}{0.18\textwidth}
\centering
\small{\textbf{Brown} cow, \textbf{dirt}.}
\end{minipage}
\;\;\begin{minipage}{0.195\textwidth}
\includegraphics[width=1\linewidth, height=1\linewidth]{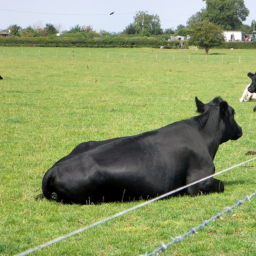}
\end{minipage}
\noindent\begin{minipage}{0.195\textwidth}
\includegraphics[width=1\linewidth, height=1\linewidth]{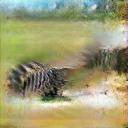}
\end{minipage}
\noindent\begin{minipage}{0.195\textwidth}
\includegraphics[width=1\linewidth, height=1\linewidth]{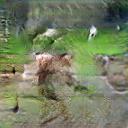}
\end{minipage}
\noindent\begin{minipage}{0.195\textwidth}
\includegraphics[width=1\linewidth, height=1\linewidth]{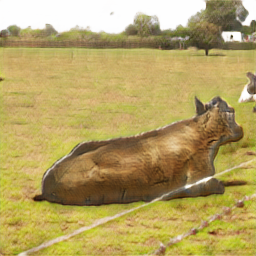}
\end{minipage}
\end{minipage}
\smallskip

\begin{minipage}{1\textwidth}
\begin{minipage}{0.18\textwidth}
\centering
\small{Boy, \textbf{road}.}
\end{minipage}
\;\;\begin{minipage}{0.195\textwidth}
\includegraphics[width=1\linewidth, height=1\linewidth]{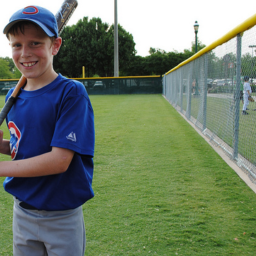}
\end{minipage}
\noindent\begin{minipage}{0.195\textwidth}
\includegraphics[width=1\linewidth, height=1\linewidth]{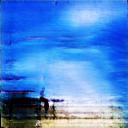}
\end{minipage}
\noindent\begin{minipage}{0.195\textwidth}
\includegraphics[width=1\linewidth, height=1\linewidth]{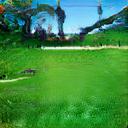}
\end{minipage}
\noindent\begin{minipage}{0.195\textwidth}
\includegraphics[width=1\linewidth, height=1\linewidth]{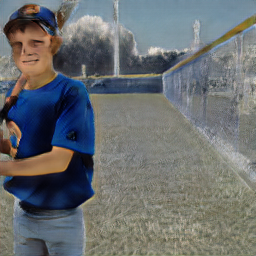}
\end{minipage}
\end{minipage}
\smallskip

\begin{minipage}{1\textwidth}
\begin{minipage}{0.195\textwidth}
\centering
\small{Given Text}
\end{minipage}
\begin{minipage}{0.195\textwidth}
\centering
\small{Original}
\end{minipage}
\noindent\begin{minipage}{0.195\textwidth}
\centering
\small{SISGAN \cite{dong2017semantic}}
\end{minipage}
\noindent\begin{minipage}{0.195\textwidth}
\centering
\small{TAGAN \cite{nam2018text}}
\end{minipage}
\noindent\begin{minipage}{0.195\textwidth}
\centering
\small{Ours}
\end{minipage}
\end{minipage}

\centering
\caption{Additional comparison results between ManiGAN, SISGAN, and TAGAN on the COCO dataset.}
\label{fig:qual_show3}
\end{figure*}

\begin{figure*}[h!]
\begin{minipage}{1\textwidth}
\begin{minipage}{0.18\textwidth}
\centering
\small{Zebra, \textbf{grass}.}
\end{minipage}
\;\;\begin{minipage}{0.195\textwidth}
\includegraphics[width=1\linewidth, height=1\linewidth]{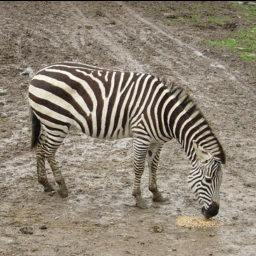}
\end{minipage}
\noindent\begin{minipage}{0.195\textwidth}
\includegraphics[width=1\linewidth, height=1\linewidth]{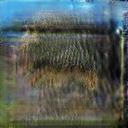}
\end{minipage}
\noindent\begin{minipage}{0.195\textwidth}
\includegraphics[width=1\linewidth, height=1\linewidth]{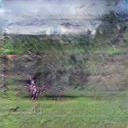}
\end{minipage}
\noindent\begin{minipage}{0.195\textwidth}
\includegraphics[width=1\linewidth, height=1\linewidth]{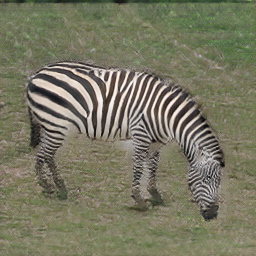}
\end{minipage}
\end{minipage}
\smallskip

\begin{minipage}{1\textwidth}
\begin{minipage}{0.18\textwidth}
\centering
\small{\textbf{Orange} bus.}
\end{minipage}
\;\;\begin{minipage}{0.195\textwidth}
\includegraphics[width=1\linewidth, height=1\linewidth]{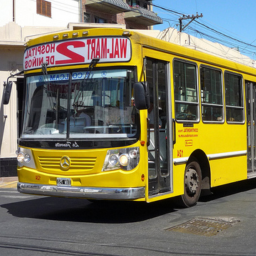}
\end{minipage}
\noindent\begin{minipage}{0.195\textwidth}
\includegraphics[width=1\linewidth, height=1\linewidth]{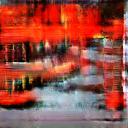}
\end{minipage}
\noindent\begin{minipage}{0.195\textwidth}
\includegraphics[width=1\linewidth, height=1\linewidth]{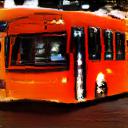}
\end{minipage}
\noindent\begin{minipage}{0.195\textwidth}
\includegraphics[width=1\linewidth, height=1\linewidth]{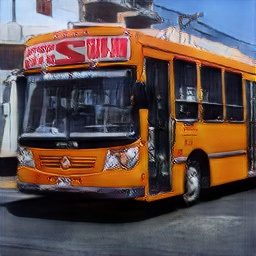}
\end{minipage}
\end{minipage}
\smallskip

\begin{minipage}{1\textwidth}
\begin{minipage}{0.18\textwidth}
\centering
\small{\textbf{Night}.}
\end{minipage}
\;\;\begin{minipage}{0.195\textwidth}
\includegraphics[width=1\linewidth, height=1\linewidth]{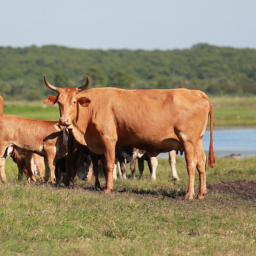}
\end{minipage}
\noindent\begin{minipage}{0.195\textwidth}
\includegraphics[width=1\linewidth, height=1\linewidth]{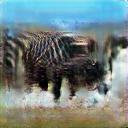}
\end{minipage}
\noindent\begin{minipage}{0.195\textwidth}
\includegraphics[width=1\linewidth, height=1\linewidth]{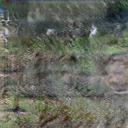}
\end{minipage}
\noindent\begin{minipage}{0.195\textwidth}
\includegraphics[width=1\linewidth, height=1\linewidth]{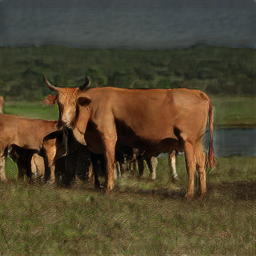}
\end{minipage}
\end{minipage}
\smallskip

\begin{minipage}{1\textwidth}
\begin{minipage}{0.18\textwidth}
\centering
\small{Kite, \textbf{green field}.}
\end{minipage}
\;\;\begin{minipage}{0.195\textwidth}
\includegraphics[width=1\linewidth, height=1\linewidth]{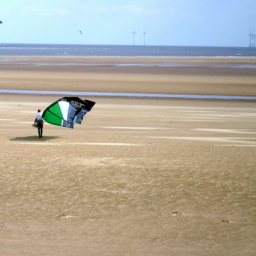}
\end{minipage}
\noindent\begin{minipage}{0.195\textwidth}
\includegraphics[width=1\linewidth, height=1\linewidth]{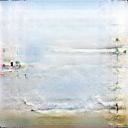}
\end{minipage}
\noindent\begin{minipage}{0.195\textwidth}
\includegraphics[width=1\linewidth, height=1\linewidth]{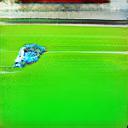}
\end{minipage}
\noindent\begin{minipage}{0.195\textwidth}
\includegraphics[width=1\linewidth, height=1\linewidth]{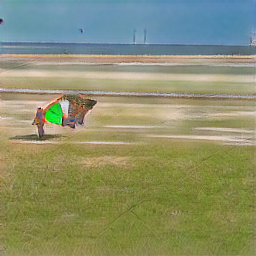}
\end{minipage}
\end{minipage}
\smallskip

\begin{minipage}{1\textwidth}
\begin{minipage}{0.18\textwidth}
\centering
\small{Pizza, \textbf{pepperoni}.}
\end{minipage}
\;\;\begin{minipage}{0.195\textwidth}
\includegraphics[width=1\linewidth, height=1\linewidth]{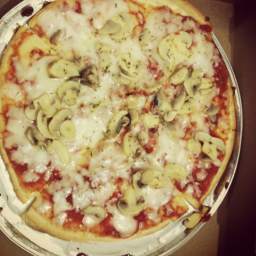}
\end{minipage}
\noindent\begin{minipage}{0.195\textwidth}
\includegraphics[width=1\linewidth, height=1\linewidth]{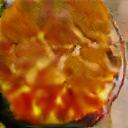}
\end{minipage}
\noindent\begin{minipage}{0.195\textwidth}
\includegraphics[width=1\linewidth, height=1\linewidth]{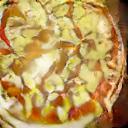}
\end{minipage}
\noindent\begin{minipage}{0.195\textwidth}
\includegraphics[width=1\linewidth, height=1\linewidth]{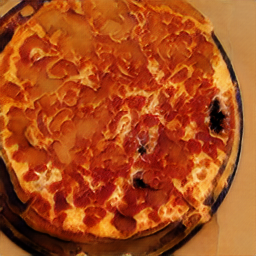}
\end{minipage}
\end{minipage}
\smallskip

\begin{minipage}{1\textwidth}
\begin{minipage}{0.18\textwidth}
\centering
\small{Zebra, \textbf{water}.}
\end{minipage}
\;\;\begin{minipage}{0.195\textwidth}
\includegraphics[width=1\linewidth, height=1\linewidth]{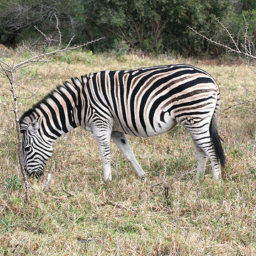}
\end{minipage}
\noindent\begin{minipage}{0.195\textwidth}
\includegraphics[width=1\linewidth, height=1\linewidth]{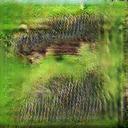}
\end{minipage}
\noindent\begin{minipage}{0.195\textwidth}
\includegraphics[width=1\linewidth, height=1\linewidth]{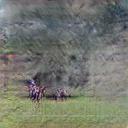}
\end{minipage}
\noindent\begin{minipage}{0.195\textwidth}
\includegraphics[width=1\linewidth, height=1\linewidth]{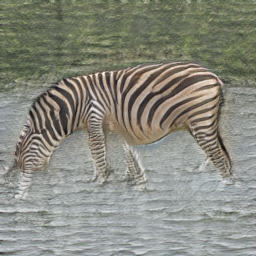}
\end{minipage}
\end{minipage}
\smallskip

\begin{minipage}{1\textwidth}
\begin{minipage}{0.195\textwidth}
\centering
\small{Given Text}
\end{minipage}
\begin{minipage}{0.195\textwidth}
\centering
\small{Original}
\end{minipage}
\noindent\begin{minipage}{0.195\textwidth}
\centering
\small{SISGAN \cite{dong2017semantic}}
\end{minipage}
\noindent\begin{minipage}{0.195\textwidth}
\centering
\small{TAGAN \cite{nam2018text}}
\end{minipage}
\noindent\begin{minipage}{0.195\textwidth}
\centering
\small{Ours}
\end{minipage}
\end{minipage}

\centering
\caption{Additional comparison results between ManiGAN, SISGAN, and TAGAN on the COCO dataset.}
\label{fig:qual_show4}
\end{figure*}

In Figs.~\ref{fig:qual_show1},~\ref{fig:qual_show2},~\ref{fig:qual_show3}, and~\ref{fig:qual_show4}, we show additional comparison results between our ManiGAN, SISGAN \cite{dong2017semantic}, and TAGAN \cite{nam2018text} on the CUB \cite{wah2011caltech} and COCO \cite{lin2014microsoft} datasets. Please watch the accompanying video for detailed comparison.


\end{document}